\documentclass[sigconf]{acmart}

\setcopyright{acmlicensed}
\copyrightyear{2018}
\acmYear{2018}
\acmDOI{XXXXXXX.XXXXXXX}
\acmConference[Conference acronym 'XX]{Make sure to enter the correct
  conference title from your rights confirmation email}{June 03--05,
  2018}{Woodstock, NY}
\acmISBN{978-1-4503-XXXX-X/2018/06}

\usepackage{xspace}
\usepackage{graphicx}
\usepackage{enumitem}
\usepackage{hyperref}
\usepackage{multirow}
\usepackage{rotating} 
\usepackage{pdflscape}

\newcommand{\red}[1]{\textcolor{red}{#1}}
\newcommand{\mat}[1]{{\bf #1}}   

\newcommand{\method}{\textsc{SVTime}\xspace}

\begin{document}

\title{\method: Small Time Series Forecasting Models Informed by ``Physics'' of Large Vision Model Forecasters}


\author{ChengAo Shen\textsuperscript{\rm 1}, Ziming Zhao\textsuperscript{\rm 1}, Hanghang Tong\textsuperscript{\rm 2}, Dongjin Song\textsuperscript{\rm 3}, Dongsheng Luo\textsuperscript{\rm 4},\\
Qingsong Wen\textsuperscript{\rm 5}, Jingchao Ni\textsuperscript{\rm 1}}
\affiliation{\textsuperscript{\rm 1}University of Houston, \textsuperscript{\rm 2}University of Illinois at Urbana-Champaign, \textsuperscript{\rm 3}University of Connecticut,\\
\textsuperscript{\rm 4}Florida International University, \textsuperscript{\rm 5}Squirrel Ai Learning\\
\textsuperscript{\rm 1}\texttt{\{cshen9,zzhao35,jni7\}@uh.edu}, \textsuperscript{\rm 2}\texttt{htong@illinois.edu}, \textsuperscript{\rm 3}\texttt{dongjin.song@uconn.edu},\\
\textsuperscript{\rm 4}\texttt{dluo@fiu.edu}, \textsuperscript{\rm 5}\texttt{qingsongedu@gmail.com}\country{}}


\renewcommand{\shortauthors}{}

\begin{abstract}

Time series AI is crucial for analyzing dynamic web content, driving a surge of pre-trained large models known for their strong knowledge encoding and transfer capabilities across diverse tasks. However, given their energy-intensive training, inference, and hardware demands, using large models as a one-fits-all solution raises serious concerns about carbon footprint and sustainability. For a specific task, a compact yet specialized, high-performing model may be more practical and affordable, especially for resource-constrained users such as small businesses. This motivates the question: Can we build cost-effective lightweight models with large-model-like performance on core tasks such as forecasting? This paper addresses this question by introducing \method, a novel \underline{S}mall model inspired by large \underline{V}ision model (LVM) forecasters for long-term \underline{Time} series forecasting (LTSF). Recently, LVMs have been shown as powerful tools for LTSF. We identify a set of key inductive biases of LVM forecasters --- analogous to the ``physics'' governing their behaviors in LTSF --- and design small models that encode these biases through 
meticulously crafted linear layers and constraint functions. Across 21 baselines spanning lightweight, complex, and pre-trained large models on 8 benchmark datasets, \method\ outperforms state-of-the-art (SOTA) lightweight models 
and rivals large models with 
$10^{3}\times$ fewer parameters than LVMs, while enabling efficient training and inference in low-resource settings.

\end{abstract}






\maketitle

\section{Introduction}\label{sec.intro}

The World Wide Web is a dynamic, ever-evolving system that continuously produce time series data pertaining to web traffic ({\em e.g.}, page views), user behavior ({\em e.g.}, bounce rates), web content ({\em e.g.}, trending topics), e-commerce ({\em e.g.}, click-through rates), system security ({\em e.g.}, latency logs), and so on, where the ability to anticipate and respond to changing patterns and user behaviors plays a crucial role. As such, time series forecasting --- analyzing historical data and predicting future trends --- emerges as an indispensible component of modern web technologies \cite{xu2021rest,hou2022multi,jhin2022exit,kamarthi2022camul,jiang2023learning}, driving intelligent web services such as content recommendation \cite{wei2023multi}, microservice monitoring \cite{jiang2023look}, and web economics modeling \cite{xu2021rest}.


\begin{figure*}[t!]
\centering
\includegraphics[width=\textwidth]{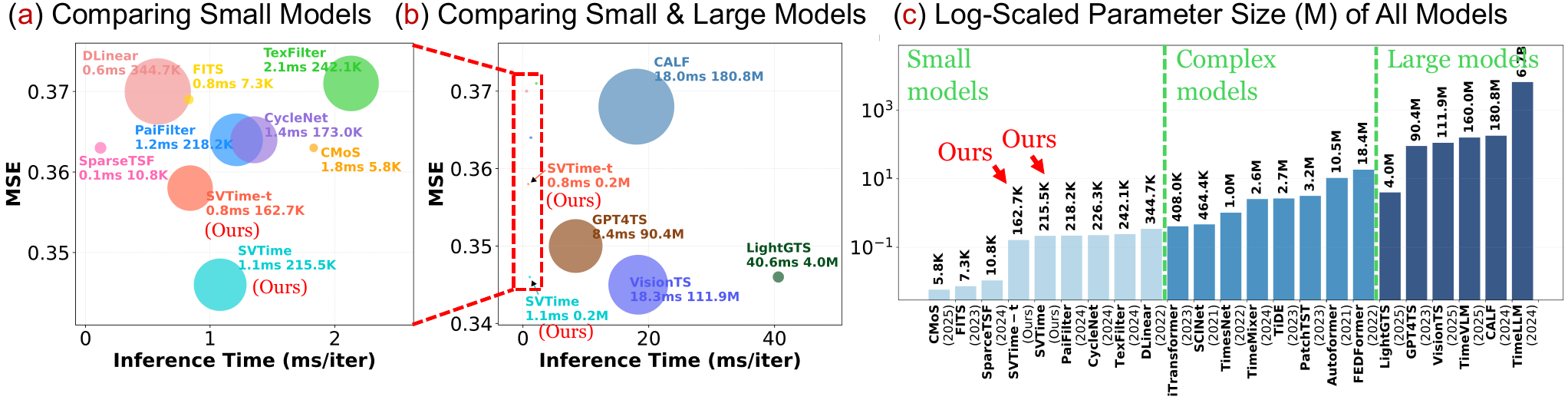}
\caption{An overview of (a) forecasting performance vs. inference time on ETTm1 dataset, where circle size reflects model size, which is a zoom-in of the small models in (b); in (b), \texttt{TimeLLM} \cite{jin2024time} is ignored for its much larger size and longer inference time (see $\S$\red{\ref{sec.exp.large_models}}); and (c) is a categorization of small, complex, and pre-trained large models.}\label{fig.intro}
\vspace{-0.2cm}
\end{figure*}

Inspired by the success of large models in AI and their strong adaptability across modalities, 
emergent methods for time series forecasting have explored Transformer \cite{zhou2021informer,wu2021autoformer,zhou2022fedformer,nie2023patchtst,liu2023itransformer}, Time Series Foundation Models (TSFMs) \cite{das2024decoder,ansari2024chronos,woo2024unified,goswami2024moment,wang2025lightgts}, Large Language Models (LLMs) \cite{zhou2023one,jin2024time,pan2024s,liu2025calf}, Large Vision Models (LVMs) \cite{chen2025visionts,shen2025visionts++,ruan2025ldm4ts} and Vision-Language Models (VLMs) \cite{zhong2025time}. However, given their energy-intensive training, inference, and hardware demands, using large models as a one-fits-all solution raises serious concerns about carbon footprint and sustainability \cite{bolon2024review}. For a specific task, recent findings reveal that 
most of a large model's parameters may be useless \cite{lyu2025occamvts}. In language domain, small language models (SLMs) such as Microsoft \texttt{Phi} series \cite{javaheripi2023phi}, NVIDIA 
\texttt{Hymba} \cite{dong2024hymba}, and \texttt{DeepSeek-R1-Distill} series \cite{guo2025deepseek} are becoming on par with LLMs on certain tasks, powering economical development of agent systems \cite{belcak2025small}. The trend toward compact yet high-performing models specialized in core tasks, along with the need for quick deployment in resource-constrained scenarios ({\em e.g.}, edge devices, small institutes and businesses), motivate the question: Can we build cost-effective models with large-model-like performance on core tasks such as long-term time series forecasting (LTSF)?


This question is challenging due to the trade-off between model capacity and efficiency. Large models --- {\em e.g.}, with millions or billions of parameters --- support complex attention mechanisms, and encoding of knowledge by large-scale pre-training. In contrast, smaller models are restricted in design possibilities and unfit to pre-training because of underfitting, thus 
appear to be less likely to rival (pre-trained) large models. A straightforward direction is to explore knowledge-distillation (KD), such as distilling \texttt{OccamVTS} from LVMs \cite{lyu2025occamvts} and \texttt{TimeDistill} from Transformers \cite{ni2025timedistill}. Whereas, KD is still resource-demanding as it relies on the 
large teacher models, which will be loaded and communicated with during the training of student models. Also, these student models 
are non-competitive in reducing parameters as evaluated in $\S$\red{\ref{sec.exp.kd}}.

Recent efforts toward small models for LTSF focus on exploring different hypotheses pertaining to the LTSF task, such as point-wise correlation \cite{zeng2023transformers}, segment-wise correlation \cite{si2025cmos}, periodicity \cite{lin2024cyclenet,lin2024sparsetsf}, and time-frequency relationships \cite{zeng2023transformers,xu2024fits}, 
while designing models with inductive biases encoding the hypotheses. For example, \texttt{SparseTSF} \cite{lin2024sparsetsf} assumes inter-period smoothness of time series and forecasts future periods by aggregating past periods. \texttt{CycleNet} \cite{lin2024cyclenet} also utilizes periods but makes them learnable in a seasonal-trend-like framework. \texttt{CMoS} \cite{si2025cmos} extends \texttt{DLinear} \cite{zeng2023transformers} and assumes the correlation between historical chunks ({\em i.e.}, a segment of time series) and future chunks can be modeled by linear layers. 
However, as the hypotheses don't align with large models, none of these models is comparable to pre-trained large models (as evaluated in $\S$\red{\ref{sec.exp}}).

In this paper, in addition to centering on the LTSF task itself, we take a new perspective by grounding our hypotheses on the behavior of large models. We analyze a specific LVM, \texttt{MAE} \cite{he2022mae} --- when used as a forecaster in \texttt{VisionTS} \cite{chen2025visionts} --- for its superior performance in LTSF over LLMs and VLMs as validated by \cite{zhao2025images,shen2025multi}. Our analysis uncovers a set of key inductive biases --- analogous to the ``physics'' governing \texttt{MAE}'s behavior in LTSF --- including (1) 
inter-period consistency ($\S$\red{\ref{sec.ib1}}); (2) 
patch-wise variety ($\S$\red{\ref{sec.ib2}}); and (3) 
distance-attenuating local attention ($\S$\red{\ref{sec.ib3}}). We design lightweight models with linear layers and constraint functions to encode these biases, sharing similar merits with physics-informed learning \cite{ji2022stden,hettige2024airphynet}, 
which is useful in learning cost-effective models. Using the first two ``physics'', we propose \method, a novel \underline{S}mall model inspired by L\underline{V}M for long-term \underline{Time} series forecasting. Including the third ``physics'' leads to a 
tiny version of the model, namely \method-t. Moreover, to complement the biases toward forecasting periods, we encapsulate our models within a backcast-residual based decomposition framework ($\S$\red{\ref{sec.backcast}}), which adaptively compensates forecasts with residual trends. As Fig. \red{\ref{fig.intro}}(a)(b) illustrate, 
with only $0.2\%$ ($0.1\%$) size of \texttt{VisionTS}, \method\ (\method-t) shows superiority in the small model regime and rivals pre-trained large models including LLMs, LVMs, TSFMs and VLMs. To sum up, our contributions are as follows.
\begin{itemize}[leftmargin=0.35cm]
\item\textbf{Discovery}. We dive in to the behavioral patterns of a SOTA LVM forecaster, uncover key inductive biases, and validate their value by explicitly transferring them to a much simpler model.
\item\textbf{Development}. We carefully design lightweight models using linear layers and constraint functions to reproduce the ``physics'' of an LVM forecaster within a parameter-limited regime.
\item\textbf{Evaluation}. We compare {\method}(-t) with 21 SOTA baselines covering lightweight, complex, and pre-trained large models on 8 benchmark datasets, along with extensive ablations, validating their small-model-like sizes and large-model-like performance.
\end{itemize}

\noindent\textbf{Categorization of Models}. In this paper, we categorize models as {\em lightweight}, {\em complex}, and {\em pre-trained large} models according to our observation of the 21 SOTA models as shown in Fig. \red{\ref{fig.intro}}(c). \texttt{DLinear} is used to separate lightweight and complex models as the smaller models mostly consist of linear layers, while the larger models employ CNN, MLP, or Transformer with more complex designs. Pre-trained large models only include models that have been pre-trained on some datasets. However, this categorization is subject to discussion and has little bearing on the essence of the paper.

\section{Related Work}

To the best of our knowledge, this is the first work to explore LVMs' inductive biases for guiding the design of lightweight forecasting models. Our work relates to \textbf{Large models for time series forecasting (TSF)}, \textbf{Lightweight models for TSF}, and \textbf{Knowledge-Distillation (KD)-based TSF}, which are discussed below.

\vspace{0.2cm}

\noindent\textbf{Large models for TSF}. Recent research on TSF draws a lot of attention to Transformer \cite{zhou2021informer,wu2021autoformer,zhou2022fedformer,nie2023patchtst,liu2023itransformer}, TSFMs \cite{das2024decoder,ansari2024chronos,woo2024unified,goswami2024moment,wang2025lightgts}, LLMs \cite{zhou2023one,jin2024time,pan2024s,liu2025calf}, LVMs \cite{chen2025visionts,shen2025visionts++,ruan2025ldm4ts}, and multimodal models \cite{zhong2025time,shen2025multi}. Early Transformer-based forecasters, such as \texttt{Informer} \cite{zhou2021informer} and \texttt{Autoformer} \cite{wu2021autoformer}, focus on encoding time points. More recent models tend to encode patches, {\em e.g.}, \texttt{PatchTST} \cite{nie2023patchtst}, or variates, {\em e.g.}, \texttt{iTransformer} \cite{liu2023itransformer}. This development inspires pre-training TSFMs on large-scale time serie datasets such as \texttt{TimesFM} \cite{das2024decoder}, \texttt{Chronos} \cite{ansari2024chronos}, \texttt{Moirai} \cite{woo2024unified}, and \texttt{LightGTS} \cite{wang2025lightgts}. The adaptation of LLMs to time series include prompt-based methods, such as \texttt{PromptCast} \cite{xue2023promptcast} and \texttt{LLMTime} \cite{gruver2023large}, and embedding-based methods, such as \texttt{GPT4TS} \cite{zhou2023one}, \texttt{TimeLLM} \cite{jin2024time}, and \texttt{CALF} \cite{liu2025calf}. More recently, LVMs such as \texttt{MAE} \cite{he2022masked} have been found more effective than LLMs by \texttt{VisionTS} \cite{chen2025visionts}, \texttt{VisionTS++} \cite{shen2025visionts++}, and the study in \cite{ruan2025ldm4ts,zhao2025images}, which further inspires multimodal models for TSF such as \texttt{TimeVLM} \cite{zhong2025time} and \texttt{DMMV} \cite{shen2025multi}. In this work, we also include some CNN-based and MLP-based models as large models for their relatively complex designs, such as \texttt{SCINet} \cite{liu2022scinet}, \texttt{TimesNet} \cite{wu2023timesnet}, and \texttt{TimeMixer} \cite{wang2024timemixer}. Despite their powerful performance, these large models are resource-demanding and may not fit resource-constrained scenarios.

\vspace{0.2cm}

\noindent\textbf{Lightweight models for TSF}. The SOTA lightweight TSF models mostly employ simple linear layers \cite{zeng2023transformers,li2023rlinear,xu2024fits,yi2024filternet,lin2024cyclenet,lin2024sparsetsf,si2025cmos}. The rationale behind their effectiveness lies in the designs for exploring certain inductive biases pertaining to the TSF task. For example, \texttt{DLinear} \cite{zeng2023transformers} assumes linear mapping between lookback window and forecasts; \texttt{CMoS} \cite{si2025cmos} assumes linear correlation between historical chunks and future chunks; \texttt{SparseTSF} \cite{lin2024sparsetsf} and \texttt{CycleNet} \cite{lin2024cyclenet} capitalize periodical patterns in TSF; while \texttt{FITS} \cite{xu2024fits} and \texttt{FilterNet} \cite{yi2024filternet} take the advantage of frequency domain for efficient modeling of temporal trends. Despite the inspiring progress, none of these models can rival the SOTA performance of large models (Fig. \red{\ref{fig.intro}}) due to their non-large-model-aligned designs. In contrast, we are exploring the possibility of developing small models with large-model-like performance.

\vspace{0.2cm}

\noindent\textbf{KD-based TSF}. A straightforward way toward small models with large-model-like performance is KD. \texttt{OccamVTS} \cite{lyu2025occamvts} is a recent cross-modal KD method that transfers TSF-essential knowledge from a pre-trained LVM teacher model to a smaller Transformer-based student model. However, as we evaluated in $\S$\red{\ref{sec.exp.kd}}, \texttt{OccamVTS}'s size still belongs to complex models. \texttt{TimeDistill} \cite{ni2025timedistill} supports cross-architecture KD, thus is more flexible in compressing student model size than \texttt{OccamVTS}. However, KD may not fit resource-constrained needs because (1) it relies on the availability (and fine-tuning) of large teacher models; (2) its training of student models involves communication with large teacher models, leading to high costs; and (3) the student model may need to be above certain size for sufficient capacity in encoding of teacher models' knowledge. As such, a standalone lightweight model may be more favorable.

\section{The Proposed \method\ Model}

\textbf{Problem Statement}. Given a multivariate time series (MTS) $\mat{X}=[\mat{x}^{1}, ..., \mat{x}^{D}]^{\top}\in\mathbb{R}^{D\times T}$ within a {\em look-back window} of length $T$, where $\mat{x}^{i}\in\mathbb{R}^{T}$ ($1\le i \le D)$ is a univariate time series (UTS) of the $i$-th variate, the goal of LTSF is to estimate the most likely values of the MTS at future $H$ time steps, {\em i.e.}, $\mat{\hat{Y}}\in\mathbb{R}^{D\times H}$, such that the difference between the estimation and the ground truth $\mat{Y}=\mat{X}_{T+1:T+H}\in\mathbb{R}^{D\times H}$ is minimized in terms of mean squared error (MSE), {\em i.e.}, $\frac{1}{D\cdot H}\sum_{i=1}^{D}\sum_{t=1}^{H}\|\mat{\hat{Y}}_{it} - \mat{Y}_{it}\|_{2}^{2}$ is minimized.

In the following, we introduce the key inductive biases (\textbf{IBs}) identified from LVM's forecasting behaviors, including (IB1) 
inter-period consistency ($\S$\red{\ref{sec.ib1}}); (IB2) 
patch-wise variety ($\S$\red{\ref{sec.ib2}}); and (IB3) distance-attenuating local attention ($\S$\red{\ref{sec.ib3}}), meanwhile reprogramming them using linear layers and constraint functions, progressively constructing {\method}(-t) models. Finally, we encapsulate our models within a lightweight backcast-residual decomposition framework for avoiding overly dominant bias toward periods ($\S$\red{\ref{sec.backcast}}). Fig. \red{\ref{fig.method}} illustrates the overall framework of {\method}(-t).

\subsection{IB1: Inter-Period Consistency}\label{sec.ib1}

\noindent{Masked} autoencoder (\texttt{MAE}) \cite{he2022masked} is pre-trained self-supervisedly by reconstructing masked image patches using ImageNet dataset. To adapt it to LTSF, \texttt{VisionTS} \cite{chen2025visionts} adopts a period-based imaging technique introduced by \texttt{TimesNet} \cite{wu2023timesnet}. Specifically, each length-$T$ UTS $\mat{x}^{i}$ is segmented into $\lfloor T/P\rfloor$ subsequences of length $P$, where $P$ is set to be the period of $\mat{x}^{i}$, which can be obtained using Fast Fourier Transform (FFT) on $\mat{x}^{i}$ \cite{wu2023timesnet} or from prior knowledge on sampling frequency. The subsequences are stacked to form a 2D image $\mat{I}^{i}\in\mathbb{R}^{P\times\lfloor T/P\rfloor}$. After standard-deviation normalization, $\mat{I}^{i}$ is duplicated 3 times to form 
an image of size $P\times\lfloor T/P\rfloor\times 3$, followed by a bilinear interpolation to resize it to an image $\mat{\tilde{I}}^{i}$ 
of size $224\times 224\times 3$ to fit the input requirement of \texttt{MAE}. 

\begin{figure}[t!]
\centering
\includegraphics[width=\columnwidth]{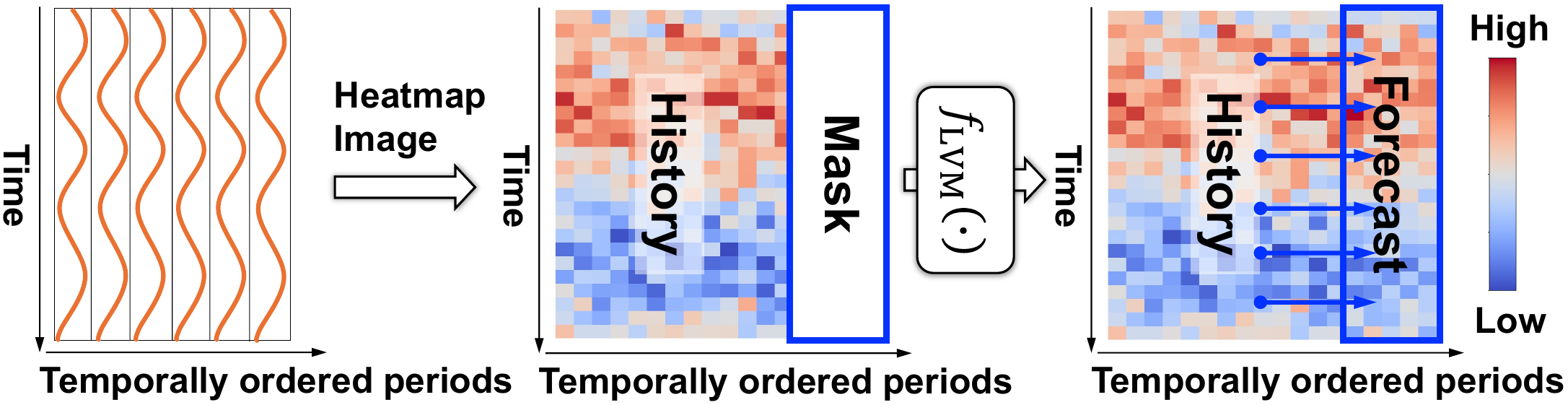}
\caption{An illustration of inter-period consistency.}\label{fig.ib1}
\end{figure}

As Fig. \red{\ref{fig.ib1}} shows, the forecast 
is achieved by reconstructing a right-appended masked area of $\mat{\tilde{I}}^{i}$, which corresponds to the future horizon of $\mat{x}^{i}$. The forecast $\mat{\hat{y}}^{i}\in\mathbb{R}^{H}$ can be recovered from the reconstructed area by de-normalization and reverse transformation. The forecast of MTS $\mat{X}$ is achieved by forecasting over $\mat{x}^{1}$, ..., $\mat{x}^{D}$ in parallel, following the channel-independence assumption \cite{nie2023patchtst}.

Due to the period-based imaging and the spatial consistency enforced during \texttt{MAE}'s pixel inference, \texttt{VitionTS} exhibits a strong bias toward \textbf{inter-period consistency} --- smoothness of values over the same within-period time point ({\em i.e.}, rows) across periods ({\em i.e.}, columns) in the image, as revealed by \cite{zhao2025images,shen2025multi}.

\subsubsection{IB1-Informed Model Design}\label{sec.ib1method}

By designing a new small model, we don't need to perform bilinear interpolation and channel duplication. Instead, we use the 2D image $\mat{I}^{i}\in\mathbb{R}^{P\times\lfloor T/P\rfloor}$ as the input ({\em i.e.} the imaged look-back window) for the $i$-the variate, and output $\mat{\hat{I}}^{i}\in\mathbb{R}^{P\times\lfloor H/P\rfloor}$ as the forecasts for $H$ time steps, where the $j$-th column $\mat{\hat{I}}_{j}^{i}\in\mathbb{R}^{P}$ ($1\le j\le\lfloor H/P\rfloor$) is the $j$-th forecasted period. $\mat{\hat{I}}^{i}$ corresponds to the masked area in Fig. \red{\ref{fig.ib1}}.

To harness inter-period consistency, let $N=\lfloor T/P\rfloor$ and $M=\lfloor H/P\rfloor$. For each column $\mat{\hat{I}}_{j}^{i}$, we introduce a set of weights $\mat{w}_{j} = [w_{j,1}, w_{j,2}, ..., w_{j,N}]$ and predict $\mat{\hat{I}}_{j}^{i}$ as a linear combination of the historical periods in $\mat{I}^{i}$, {\em i.e.}, $\mat{\hat{I}}_{j}^{i} = \mat{I}^{i}\mat{w}_{j}^{\top}$. Let $\mat{W} = [\mat{w}_{1}^{\top}, ..., \mat{w}_{M}^{\top}]\in\mathbb{R}^{N\times M}$, the forecast of $M$ periods is $\mat{\hat{I}}^{i} = \mat{I}^{i}\mat{W}\in\mathbb{R}^{P\times M}$. It is noteworthy that $\mat{W}$ is shared across variates. Then the forecasted time series $\mat{\hat{y}}^{i}\in\mathbb{R}^{H}$ can be recovered from $\mat{\hat{I}}^{i}$ in a similar way as aforementioned, and multiple variates for $1\le i \le D$ in $\mat{X}$ can be forecasted in parallel according to the channel-independence assumption \cite{nie2023patchtst}.

\subsection{IB2: Patch-Wise Variety}\label{sec.ib2}

\begin{figure}[h!]
\centering
\includegraphics[width=0.9\columnwidth]{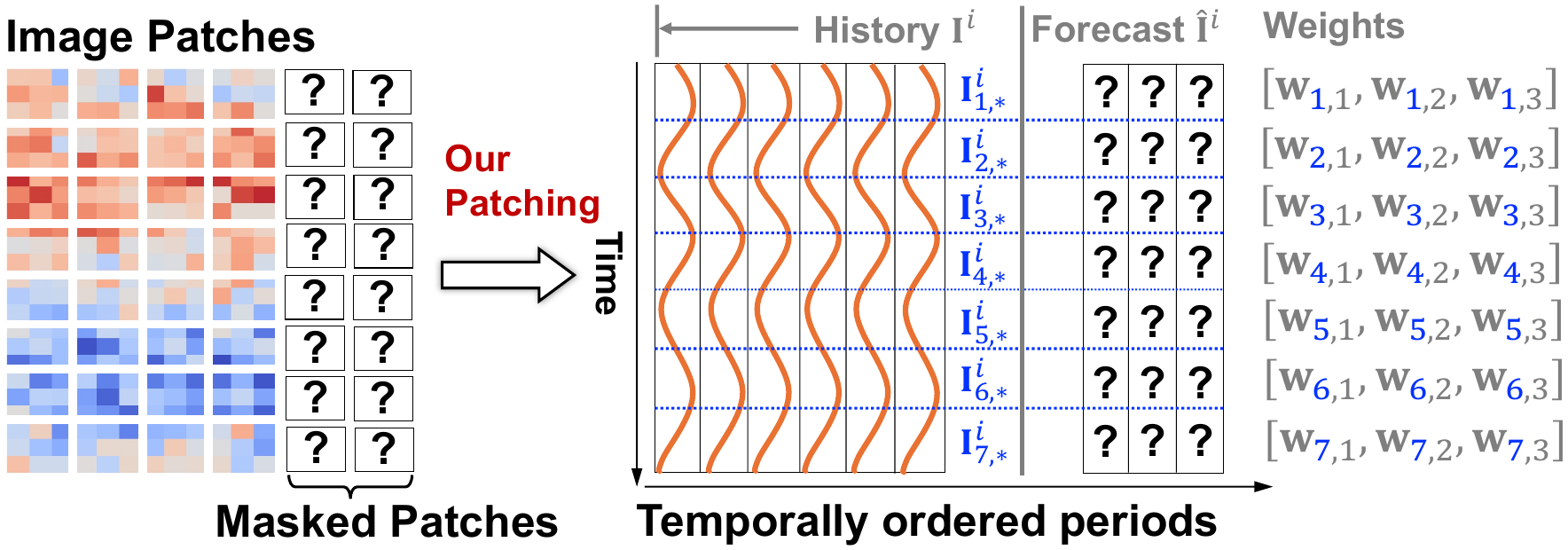}
\caption{An illustration of patch-wise variety.}\label{fig.ib2}
\end{figure}

\noindent{As} Fig. \red{\ref{fig.ib2}} illustrates, using $\texttt{ViT}$ backbone \cite{dosovitskiy2020vit}, $\texttt{MAE}$ divides an input image into a fixed number of patches, and encodes them for reconstructing the patches in the masked area. This mechanism enables \textbf{patch-wise variety} --- each row of patches in the image may have its row-specific inter-period consistency, while the degree of consistency may vary across different rows of patches.

\subsubsection{IB2-Informed Model Design}\label{sec.ib2method}

To harness IB2, we divide each period in the historical image $\mat{I}^{i}$ into $K$ patches, each is of length-$\lfloor P/K\rfloor$ (additional time points will be allocated to the last patch), as illustrated in Fig. \red{\ref{fig.ib2}} (where $K=7$). Then the forecasting of $\mat{\hat{I}}^{i}$ will be performed in patch-wise.

Let the $k$-th row of the patches in $\mat{I}^{i}$ as $\mat{I}_{k,*}^{i} \in \mathbb{R}^{\lfloor P/K\rfloor\times N}$, and let the $k$-th patch in the $j$-th column of $\mat{\hat{I}}^{i}$ as $\mat{\hat{I}}_{k,j}^{i}\in\mathbb{R}^{\lfloor P/K\rfloor}$ ($j=1,2,3$ in Fig. \red{\ref{fig.ib2}}), we introduce weights $\mat{w}_{k,j} = [w_{k,j,1}, ..., w_{k,j,N}]$ and predict $\mat{\hat{I}}_{k,j}^{i}$ as a linear combination of the historical patches in $\mat{I}_{k,*}^{i}$, {\em i.e.}, $\mat{\hat{I}}_{k,j}^{i} = \mat{I}_{k,*}^{i}\mat{w}_{k,j}^{\top}$. Let $\mat{W}_{k} = [\mat{w}_{k,1}^{\top}, ..., \mat{w}_{k,M}^{\top}]\in\mathbb{R}^{N\times M}$, the forecast of the $k$-th patches for all of the $M$ future periods becomes $\mat{\hat{I}}_{k,*}^{i}=\mat{I}_{k,*}^{i}\mat{W}_{k}\in\mathbb{R}^{\lfloor P/K\rfloor\times M}$. Forecasting $\mat{\hat{I}}_{k,*}^{i}$ for $1\le k \le K$ in parallel accomplishes the forecasting of $\mat{\hat{I}}^{i}$, from which we can recover the forecasted time series $\mat{\hat{y}}\in\mathbb{R}^{H}$ as before.

Comparing with the model in $\S$\red{\ref{sec.ib1method}}, the model in this section extends a single $\mat{W}$ to $\mat{W}_{1}$, ..., $\mat{W}_{K}$ for fine-grained forecasting. As we evaluated in the ablation $\S$\red{\ref{sec.exp.ablation}}, this extension brings substantial performance improvement. 

\vspace{0.2cm}

\noindent\textbf{Remark}. Our notion of ``patch'' refers specifically to {\em within-period patches}, and is used for inter-period consistency. It is different from the ``patch'' used in existing methods such as \texttt{PatchTST} \cite{nie2023patchtst}, where a patch is a segment selected without using period.

\subsection{IB3: Distance-Attenuating Local Attention}\label{sec.ib3}

\begin{figure}[h!]
\centering
\includegraphics[width=0.95\columnwidth]{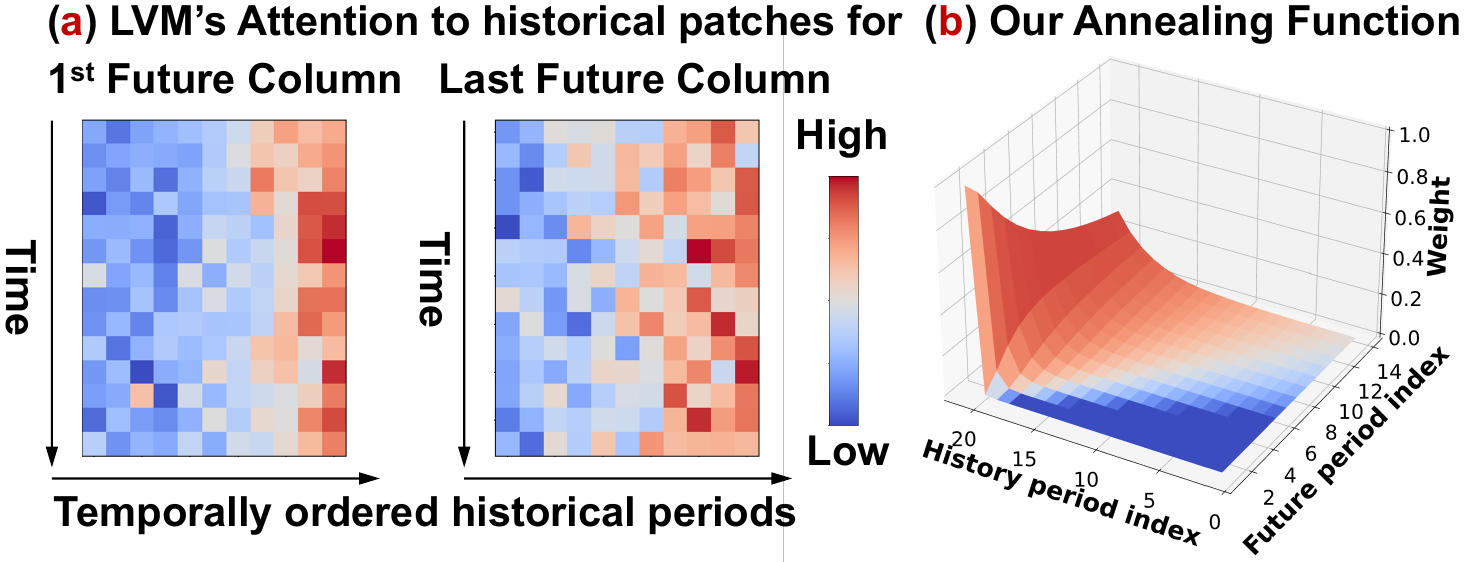}
\caption{An illustration of (a) distance-attenuating local attention; and (b) our annealing constraint function.}\label{fig.ib3}
\end{figure}

\noindent{We} also investigate the attention scores of \texttt{MAE} when forecasting different future periods ({\em i.e.}, different columns in the masked area of Fig \red{\ref{fig.ib1}}). Fig. \red{\ref{fig.ib3}}(a) shows the attention scores to the history for the first and last columns in the masked area. As can be seen, when forecasting the first period, the model focuses on the last several periods in the history, which are nearby to it. This {\em local attention}, however, {\em attenuates} when forecasting the last, {\em distant} period, where the model attention becomes more uniform across the entire history. This \textbf{distance-attenuating local attention} is reasonable since short-term forecasts may rely more on local patterns while long-term forecasts may depend on global patterns.

\begin{figure*}[t!]
\centering
\includegraphics[width=0.75\textwidth]{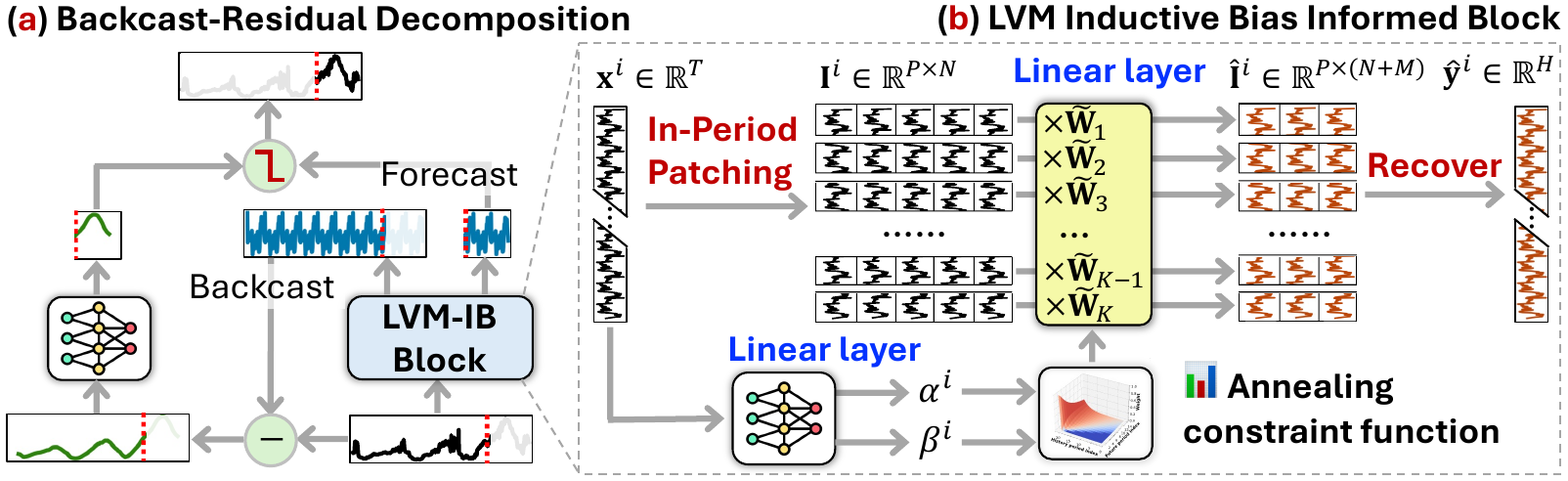}
\caption{An overview of \method\ framework. (a) {\method}(-t) uses a backcast-residual decomposition to adaptively learn trend and seasonal components. (b) LVM-IB block uses linear layers and a constraint function to encode IB1-IB3 ($\S$\red{\ref{sec.ib1}}-$\S$\red{\ref{sec.ib3}}).}
\label{fig.method}
\end{figure*}

\subsubsection{IB3-Informed Model Design}\label{sec.ib3method}

To encode this IB in our model, we propose to replace the weights $\mat{W}_{1}$, ..., $\mat{W}_{K}$ by a novel {\em annealing constraint function} powered weighting mechanism. For the $j$-th period in the forecast $\mat{\hat{I}}^{i}$, the constraint function should generate weights $w_{j,1}$, ..., $w_{j,N}$ such that (1) when $j$ is small ({\em e.g.}, close to 1), the weights focus on local areas, {\em i.e.}, weights close to $w_{j,N}$ are remarkably larger than those close to $w_{j,1}$; and (2) when $j$ is large ({\em e.g.}, close to M), the weights tend to be uniform, {\em i.e.}, the differences of weights between $w_{j,1}$ and $w_{j,N}$ is small. To this end, we propose the following annealing function $w(j, n)$:
\begin{equation}\label{eq.constraint}
\begin{aligned}
w(j, n) = \frac{\tilde{w}(j,n)}{\sum_{n=1}^{N}\tilde{w}(j,n)},~~\text{with}~\tilde{w}(j, n)=\exp\left(\frac{\alpha\cdot(n - N)}{1 + \beta\cdot(j - 1)}\right)
\end{aligned}
\end{equation}
where the numerator of $\tilde{w}(j,n)$ ensures $\tilde{w}(j,N)$ is larger than $\tilde{w}(j,1)$ for a fixed $j$ ({\em i.e.}, local attention). $\alpha$ is a parameter to scale the difference when $n$ changes. The denominator of $\tilde{w}(j,n)$ smooths the weights when $j$ is large. $\beta$ is a parameter to control the degree of smoothness. $\exp(\cdot)$ ensures positive weights and the normalization in $w(j,n)$ makes the weights sum up to 1.

Fig. \red{\ref{fig.ib3}}(b) shows the synergy between the numerator and denominator in Eq. (\red{\ref{eq.constraint}}), where short-term future periods have larger weights on nearby historical periods (N=20), while long-term future periods have more uniform weights, satisfying our design criteria.

Moreover, instead of manually tuning $\alpha$ and $\beta$ as hyperparameters in Eq. (\red{\ref{eq.constraint}}), we introduce linear layers to learn each of them from the input time series $\mat{x}^{i}\in\mathbb{R}^{T}$ automatically:
\begin{equation}\label{eq.scaler}
\begin{aligned}
\alpha^{i} = \text{SoftPlus}((\mat{w}^{\alpha})^{\top}\mat{x}^{i} + b^{\alpha}),~~~\beta^{i} = \text{SoftPlus}((\mat{w}^{\beta})^{\top}\mat{x}^{i} + b^{\beta})
\end{aligned}
\end{equation}
where $\mat{w}^{\alpha},\mat{w}^{\beta}\in\mathbb{R}^{T}$, $b^{\alpha}$ and $b^{\beta}$ are parameters of the two linear layers. $\text{SoftPlus}(\cdot)$ is applied to avoid negative scaling. Eq. (\red{\ref{eq.scaler}}) enables different $\alpha^{i}$'s ($\beta^{i}$'s) for different variates $\mat{x}^{i}$ ($1\le i\le D)$.

Finally, following IB2 in $\S$\red{\ref{sec.ib2}}, we apply Eq. (\red{\ref{eq.constraint}}) to patches within periods. To this end, let $\mat{\tilde{W}}=[w(j,n)]_{1\le j\le M,1\le n\le N}^{\top}\in\mathbb{R}^{N\times M}$ be the collection of weights from Eq. (\red{\ref{eq.constraint}}), we introduce learnable scalar weights $w_{1}^{p}$, ..., $w_{K}^{p}$ for $K$ patches, and define $\mat{\tilde{W}}_{k}=w_{k}^{p}\mat{\tilde{W}}$ where $1\le k\le K$. Then the set of weights $\mat{\tilde{W}}_{1}$, ..., $\mat{\tilde{W}}_{K}$, which encode our constraint function, replaces the fully learnable weights $\mat{W}_{1}$, ... $\mat{W}_{K}$ in $\S$\red{\ref{sec.ib2method}}, analogous to physics-informed learning. It is noteworthy that, this weighting mechanism only uses $O(T+K)$ parameters, in contrast to the $O(KMN)$ parameters in $\S$\red{\ref{sec.ib2method}}.

\subsection{Backcast-Residual Decomposition}\label{sec.backcast}

From IB1 ($\S$\red{\ref{sec.ib1}}), LVMs tend to forecast periodic patterns while may overlook the global trend, as validated by \cite{zhao2025images}. To complement this bias, 
we encapsulate our model within a lightweight backcast-residual decomposition framework \cite{shen2025multi} as shown in Fig. \red{\ref{fig.method}}(a).

Let $\texttt{LVM-IB}(\cdot)$ be the proposed block that encode IB1-IB3 ($\S$\red{\ref{sec.ib1}}-$\S$\red{\ref{sec.ib3}}). In addition to forecast $\mat{\hat{y}}^{i}$, the decomposition framework applies $\texttt{LVM-IB}(\cdot)$ to ``backcast'' the lookback window $\mat{x}^{i}$, {\em i.e.}, $[\mat{\hat{x}}^{i}, \mat{\hat{y}}^{i}] = \texttt{LVM-IB}(\mat{x}^{i})$ (the details of backcast is deferred to $\S$\red{\ref{sec.summary}}). Due to the period-prone prediction, $\mat{\hat{y}}^{i}$ tends to be forecasted periods, $\mat{\hat{x}}^{i}$ tend to be the seasonal component of $\mat{x}^{i}$
, and the residual $\Delta\mat{x}^{i}=\mat{x}^{i}-\mat{\hat{x}}^{i}$ tend to be the trend component. Thus, $\Delta\mat{x}^{i}$ is fed to a linear layer to forecast the trend component $\Delta\mat{\hat{y}}^{i}\in\mathbb{R}^{H}$. Finally, $\Delta\mat{\hat{y}}^{i}$ is combined with the period forecast $\mat{\hat{y}}^{i}$ to determine the final forecast $\mat{\hat{y}}_{\text{final}}^{i}$. The overall process can be summarized as:
\begin{equation}\label{eq.backcast}
\begin{aligned}
&[\mat{\hat{x}}^{i}, \mat{\hat{y}}^{i}] = \texttt{LVM-IB}(\mat{x}^{i})\rightarrow\Delta\mat{x}^{i} = \mat{x}^{i} - \mat{\hat{x}}^{i}\rightarrow\\
&\Delta\mat{y}^{i} = \Delta\mat{x}^{i}\mat{W}^{B} + \mat{b}^{B}\rightarrow\mat{\hat{y}}_{\text{final}}^{i} = g\circ\Delta\mat{y}^{i} + (1-g)\circ\mat{\hat{y}}^{i}
\end{aligned}
\end{equation}
where $\mat{W}^{B}\in\mathbb{R}^{T\times H}$, $\mat{b}^{B}\in\mathbb{R}^{H}$ are parameters of the linear layer for trend forecasting, and $g=\texttt{sigmoid}(w^{g})\in[0,1]$ is a lightweight gate with a learnable scalar parameter $w^{g}$.

By doing so, the proposed {\method}(-t) can compensate the strong bias toward forecasting periods.

\subsection{Summary of {\method}(-t) Model}\label{sec.summary}

Putting all components together, Fig. \red{\ref{fig.method}} summarizes the overall framework, 
where Fig. \red{\ref{fig.method}}(b) shows the $\texttt{LVM-IB}(\cdot)$ block. In this paper, we study two instantiations of the $\texttt{LVM-IB}(\cdot)$ block, leading to two overall models: (1) \textbf{\method} uses only IB1 and IB2 and sets $\mat{W}_{1}$, ..., $\mat{W}_{K}$ as learnable parameters ({\em i.e.}, the model in $\S$\red{\ref{sec.ib2method}}); and (2) \textbf{\method-t} uses IB1-IB3 and configures $\mat{\tilde{W}}_{1}$, ..., $\mat{\tilde{W}}_{K}$ with annealing constraint function ({\em i.e.}, the model in $\S$\red{\ref{sec.ib3method}}).

They have different mechanisms for backcast in Eq. (\red{\ref{eq.backcast}}): (1) \textbf{\method} simply extends $\mat{W}_{k}$ from an $N$-by-$M$ matrix to an $N$-by-($N+M$) matrix for all $1\le k \le K$, {\em i.e.}, predicting $N$ more periods in $\mat{\hat{x}}^{i}$ to represent an reconstruction of $\mat{x}^{i}$; (2) \textbf{\method-t} uses $\mat{\tilde{W}}_{k}$ ($1\le 1\le K)$ to forecast $\mat{\hat{y}}^{i}$ while uses the learnable scalar $[w_{1}^{p}, ..., w_{K}^{p}]$ only (without constraint function) to backcast $\mat{\hat{x}}^{i}$. 

Additionally, our $\texttt{LVM-IB}(\cdot)$ block allows stacking multiple blocks by properly setting input/output dimensions. For example,
\begin{equation}
\begin{aligned}
[\mat{\hat{x}}_{l+2}^{i}, \mat{\hat{y}}^{i}] = \texttt{LVM-IB}(\mat{\hat{x}}_{l+1}^{i}),~~~\text{where}~\mat{\hat{x}}_{l+1}^{i} = \texttt{LVM-IB}(\mat{\hat{x}}_{l}^{i}),
\end{aligned}
\end{equation}
where $\mat{\hat{x}}_{0}^{i}=\mat{x}^{i}$. This allows re-parameterizing the history as $\mat{\hat{x}}_{l+1}^{i}$ before performing the final backcasting and forecasting. Empirically, this trick is found useful in improving forecasting performance.

\vspace{0.2cm}

\noindent\textbf{Training}. After obtaining $\mat{\hat{Y}}=[\mat{\hat{y}}^{1}, ..., \mat{\hat{y}}^{D}]$, {\method}(-t) is trained by minimizing MSE, {\em i.e.}, $\frac{1}{D\cdot H}\sum_{i=1}^{D}\sum_{t=1}^{H}\|\mat{\hat{Y}}_{it} - \mat{Y}_{it}\|_{2}^{2}$.

\vspace{0.2cm}

\noindent\textbf{Complexity}. The parameter size of a single $\texttt{LVM-IB}(\cdot)$ block in \method\ is $O(KN(M+N)+TH)$, and that of \method-t is $O(T + K + TH)$, which further reduces the size. Since $K$, $M$, $N$ are small, the main overhead comes from $TH$ --- the linear layer in Eq. (\red{\ref{eq.backcast}}). However, as shown in Fig. \red{\ref{fig.intro}}, these sizes are competitively small, especially considering the earned performance gains.

\section{Experiments}\label{sec.exp}


\noindent\textbf{Datasets}. We adopt 8 widely used MTS benchmarks: ETT (Electricity Transformer Temperature), including ETTh1, ETTh2, ETTm1, ETTm2 \cite{zhou2021informer}; Weather \cite{zhou2021informer}; Electricity \cite{zhou2021informer}, Traffic \cite{wu2021autoformer}; and Solar-Energy \cite{liu2023itransformer}. Following standard protocols \cite{wu2021autoformer,nie2023patchtst}, we split the datasets chronologically into training/validation/test sets using a $60\%/20\%/20\%$ ratio for ETT and $70\%/10\%/20\%$ for the others. The prediction horizon H is set to $\{96,192,336,720\}$ for all datasets. The look-back $T$ is fixed at 512. 
The details about all of the datasets can be found in Appendix \red{\ref{app.Datasets}}.

\vspace{0.2cm}

\noindent\textbf{The Compared Methods}. We compare {\method}(-t) with 21 SOTA methods, including seven \textbf{lightweight models}: (1) \texttt{DLinear} \cite{zeng2023transformers}; (2) \texttt{FITS} \cite{xu2024fits}, (3) \texttt{TexFilter} \cite{yi2024filternet}; (4) \texttt{PaiFilter} \cite{yi2024filternet}; (5) \texttt{SparseTSF} \cite{lin2024sparsetsf}; (6) \texttt{CycleNet} \cite{lin2024cyclenet}; (7) \texttt{CMoS} \cite{si2025cmos}; eight \textbf{complex models}: (8) \texttt{TimeMixer} \cite{wang2024timemixer}; (9) \texttt{TiDE} \cite{das2023tide}; (10) \texttt{SCINet} \cite{liu2022scinet}; (11) \texttt{TimesNet} \cite{wu2023timesnet}; (12) \texttt{iTransformer} \cite{liu2023itransformer}; (13) \texttt{PatchTST} \cite{nie2023patchtst}; (14) \texttt{FEDFormer} \cite{zhou2022fedformer}; (15) \texttt{Autoformer} \cite{wu2021autoformer}; six \textbf{pre-trained large models} covering \textbf{LVM}: (16) \texttt{VisionTS} \cite{chen2025visionts}; \textbf{LLMs}: (17) \texttt{GPT4TS} \cite{zhou2023one}; (18) \texttt{TimeLLM} \cite{jin2024time}; (19) \texttt{CALF} \cite{liu2025calf}; \textbf{VLM}: (20) \texttt{TimeVLM} \cite{zhong2025time}; and \textbf{TSFM}: (21) \texttt{LightGTS} \cite{wang2025lightgts}. Here, \texttt{LightGTS} is selected for its better performance than other SOTA TSFMs such as \texttt{Moirai} \cite{liu2025moirai}, \texttt{Chronos} \cite{ansari2024chronos}, and \texttt{Time-MoE} \cite{shi2025timemoe}, as reported in \cite{wang2025lightgts}.

Additionally, we compare our method with a KD method --- \texttt{OccamVTS} \cite{lyu2025occamvts} --- specifically in $\S$\red{\ref{sec.exp.analysis}}. For our method, we evaluate both \method\ and \method-t. The hyperparameter $K$ --- the number of patches ($\S$\red{\ref{sec.ib2method}}) --- is set as $\lfloor P/6\rfloor$. We analyze the impact of $K$ in $\S$\red{\ref{sec.exp.analysis}}. The number of $\texttt{LVM-IB}(\cdot)$ block is searched within $[1, 3]$ using validation set. Ablation studies include several variants of our methods ($\S$\red{\ref{sec.exp.ablation}}). Following \cite{chen2025visionts}, the imaging period $P$ for \texttt{VisionTS}, \method, and \method-t is set based on each dataset's sampling frequency (see Appendix \red{\ref{app.Datasets}}).


\vspace{0.2cm}

\noindent{\textbf{Evaluation}}. Following \cite{nie2023patchtst,zeng2023transformers,tan2024language}, we use Mean Squared Error (MSE) and Mean Absolute Error (MAE) to evaluate the LTSF performance of the compared methods, and use parameter size, GPU cost, training/inference time, {\em etc.}, to evaluate model complexity.


\subsection{Experiment Results}
\begin{table*}[htb]
\centering
\caption{LTSF performance comparison of lightweight models on benchmark datasets. The results are averaged over 5 runs across prediction horizons $H \in\{96, 192, 336, 720\}$. Lower MSE and MAE indicate better performance. \red{Red} (\textcolor{blue}{\underline{blue}}) values indicate the best (second-best) MSE and MAE per row. Full results are available in Appendix \red{\ref{app.FullResults}}.
} \label{tab.light weight}
\resizebox{\textwidth}{!}{
\begin{tabular}{l|cc|cc|cc|cc|cc|cc|cc|cc|c}
\toprule[1pt]
                     & \multicolumn{2}{c}{ETTh1}                                   & \multicolumn{2}{c}{ETTh2}                                   & \multicolumn{2}{c}{ETTm1}                                   & \multicolumn{2}{c}{ETTm2}                                   & \multicolumn{2}{c}{Weather}                                 & \multicolumn{2}{c}{Electricity}                             & \multicolumn{2}{c}{Traffic}                                 & \multicolumn{2}{c}{Solar}                                   &           \\
        Model                    & MSE                          & MAE                          & MSE                          & MAE                          & MSE                          & MAE                          & MSE                          & MAE                          & MSE                          & MAE                          & MSE                          & MAE                          & MSE                          & MAE                          & MSE                          & MAE                          & Wins  \\\midrule
\texttt{DLinear} (2022)   &  0.447                        & 0.459                        & 0.442                        & 0.447                        & 0.370                        & 0.396                        & 0.264                        & 0.329                        & 0.245                        & 0.300                        & 0.163                        & 0.261                        & 0.421                        & 0.294                        & 0.232                        & 0.297                        & \textcolor{red}{\textbf{0}} (\textcolor{blue}{\underline{{0}}})      \\
\texttt{FITS} (2023)      &  0.430                        & 0.439                        & \textcolor{red}{\textbf{0.346}} & \textcolor{blue}{\underline{0.392}} & 0.369                        & 0.387                        & \textcolor{blue}{\underline{0.258}} & 0.317                        & 0.245                        & 0.283                        & 0.296                        & 0.400                        & 0.431                        & 0.305                        & 0.774                        & 0.705                        & \textcolor{red}{\textbf{1}} (\textcolor{blue}{\underline{{2}}})      \\
\texttt{TexFilter} (2024)      & 0.424                        & 0.446                        & 0.372                        & 0.410                        & 0.371                        & 0.399                        & 0.288                        & 0.340                        & 0.232                        & 0.270                        & 0.167                        & 0.264                        & 0.415                        & 0.299                        & \textcolor{blue}{\underline{0.209}} & 0.271                        & \textcolor{red}{\textbf{0}} (\textcolor{blue}{\underline{{1}}})      \\
\texttt{PaiFilter} (2024)     & 0.429                        & 0.442                        & 0.371                        & 0.409                        & 0.364                        & 0.390                        & 0.265                        & 0.326                        & \textcolor{red}{\textbf{0.223}} & \textcolor{red}{\textbf{0.262}} & 0.165                        & 0.259                        & 0.416                        & 0.294                        & \textcolor{red}{\textbf{0.199}} & \textcolor{blue}{\underline{0.255}} & \textcolor{red}{\textbf{3}} (\textcolor{blue}{\underline{{1}}})      \\
\texttt{SparseTSF} (2024)       & 0.425                        & 0.444                        & 0.360                        & 0.399                        & 0.363                        & 0.382                        & \textcolor{red}{\textbf{0.256}} & \textcolor{red}{\textbf{0.314}} & 0.244                        & 0.281                        & 0.197                        & 0.292                        & 0.432                        & 0.298                        & 0.237                        & 0.269                        & \textcolor{red}{\textbf{2}} (\textcolor{blue}{\underline{{0}}})      \\
\texttt{CycleNet} (2024)      & 0.419                        & 0.431                        & 0.358                        & 0.396                        & 0.364                        & 0.386                        & 0.258                        & 0.317                        & 0.241                        & 0.279                        & \textcolor{blue}{\underline{0.158}} & \textcolor{blue}{\underline{0.251}} & \textcolor{blue}{\underline{0.412}} & 0.287                        & 0.229                        & 0.289                        & \textcolor{red}{\textbf{0}} (\textcolor{blue}{\underline{{3}}})      \\
\texttt{CMoS} (2025)         & \textcolor{red}{\textbf{0.416}} & 0.431                        & 0.355                        & 0.399                        & 0.363                        & 0.384                        & 0.267                        & 0.323                        & \textcolor{blue}{\underline{0.231}} & 0.269                        & 0.165                        & 0.257                        & 0.424                        & \textcolor{blue}{\underline{0.286}} & 0.224                        & 0.263                        & \textcolor{red}{\textbf{1}} (\textcolor{blue}{\underline{{2}}})      \\\midrule
\method-t (Ours)        & \textcolor{blue}{\underline{0.417}} & \textcolor{blue}{\underline{0.430}} & 0.357                        & 0.399                        & \textcolor{blue}{\underline{0.358}} & \textcolor{blue}{\underline{0.379}} & 0.259                        & \textcolor{blue}{\underline{0.316}} & 0.231                        & \textcolor{blue}{\underline{0.269}} & 0.165                        & 0.258                        & 0.419                        & 0.286                        & 0.232                        & 0.273                        & \textcolor{red}{\textbf{0}} (\textcolor{blue}{\underline{{6}}})      \\
\method (Ours)        & 0.418                        & \textcolor{red}{\textbf{0.421}} & \textcolor{blue}{\underline{0.351}} & \textcolor{red}{\textbf{0.386}} & \textcolor{red}{\textbf{0.346}} & \textcolor{red}{\textbf{0.369}} & 0.265                        & 0.321                        & 0.240                        & 0.280                        & \textcolor{red}{\textbf{0.157}} & \textcolor{red}{\textbf{0.247}} & \textcolor{red}{\textbf{0.378}} & \textcolor{red}{\textbf{0.248}} & 0.213                        & \textcolor{red}{\textbf{0.245}} & \textcolor{red}{\textbf{9}} (\textcolor{blue}{\underline{{1}}})     \\ \bottomrule[1pt]
\end{tabular}
}
\end{table*}

\begin{table*}[htb]
\centering
\caption{LTSF perforamnce comparison of complex models on benchmark datasets. Full results are available in Appendix \red{\ref{app.FullResults}}.
}\label{tab.complex model}
\resizebox{\textwidth}{!}{
\begin{tabular}{l|l|cc|cc|cc|cc|cc|cc|cc|cc|c}
\toprule[1pt]
     &  & \multicolumn{2}{c}{ETTh1}                                   & \multicolumn{2}{c}{ETTh2}                                   & \multicolumn{2}{c}{ETTm1}                                   & \multicolumn{2}{c}{ETTm2}                                   & \multicolumn{2}{c}{Weather}                                 & \multicolumn{2}{c}{Electricity}                             & \multicolumn{2}{c}{Traffic}                                 & \multicolumn{2}{c}{Solar}                                   &       \\
    &Model                    & MSE                          & MAE                          & MSE                          & MAE                          & MSE                          & MAE                          & MSE                          & MAE                          & MSE                          & MAE                          & MSE                          & MAE                          & MSE                          & MAE                          & MSE                          & MAE                          & Wins  \\\midrule
\multirow{2}{*}{\rotatebox{90}{MLP}} &\texttt{TimeMixer} (2024)    & 0.470                        & 0.475                        & 0.353                        & 0.402                        & 0.429                        & 0.428                        & 0.312                        & 0.354                        & 0.244                        & 0.280                        & 0.185                        & 0.286                        & 0.438                        & 0.320                        & 0.225                        & 0.280                        & \textcolor{red}{\textbf{0}} (\textcolor{blue}{\underline{{0}}})      \\
&\texttt{TiDE} (2023)      & 0.421                        & 0.433                        & \textcolor{red}{\textbf{0.343}} & \textcolor{blue}{\underline{0.389}} & 0.366                        & 0.385                        & \textcolor{red}{\textbf{0.257}} & \textcolor{red}{\textbf{0.315}} & 0.243                        & 0.280                        & 0.165                        & \textcolor{blue}{\underline{0.258}} & 0.436                        & 0.313                        & 0.236                        & 0.275                        & \textcolor{red}{\textbf{3}} (\textcolor{blue}{\underline{{2}}}) \\\hline
\multirow{2}{*}{\rotatebox{90}{CNN}} &\texttt{SCINet} (2021)               & 0.483                        & 0.472                        & 0.399                        & 0.428                        & 0.427                        & 0.432                        & 0.296                        & 0.347                        & 0.259                        & 0.294                        & 0.217                        & 0.324                        & 0.510                        & 0.400                        & 0.227                        & 0.310                        & \textcolor{red}{\textbf{0}} (\textcolor{blue}{\underline{{0}}}) \\
&\texttt{TimesNet} (2022)           & 0.538                        & 0.514                        & 0.397                        & 0.434                        & 0.446                        & 0.438                        & 0.323                        & 0.358                        & 0.275                        & 0.305                        & 0.214                        & 0.311                        & 0.623                        & 0.335                        & 0.216                        & 0.287                        & \textcolor{red}{\textbf{0}} (\textcolor{blue}{\underline{{0}}}) \\\hline

\multirow{4}{*}{\rotatebox{90}{Transformer}} &\texttt{Autoformer} (2021)            & 0.544                        & 0.535                        & 0.438                        & 0.480                        & 0.569                        & 0.505                        & 0.340                        & 0.389                        & 0.355                        & 0.397                        & 0.276                        & 0.375                        & 0.666                        & 0.407                        & 0.848                        & 0.692                        & \textcolor{red}{\textbf{0}} (\textcolor{blue}{\underline{{0}}}) \\
&\texttt{FEDFormer} (2022)        & 0.480                        & 0.498                        & 0.437                        & 0.480                        & 0.432                        & 0.458                        & 0.343                        & 0.392                        & 0.366                        & 0.413                        & 0.231                        & 0.343                        & 0.610                        & 0.373                        & 0.330                        & 0.415                        & \textcolor{red}{\textbf{0}} (\textcolor{blue}{\underline{{0}}}) \\
&\texttt{PatchTST} (2023)      & 0.434                        & 0.452                        & \textcolor{blue}{\underline{0.347}} & 0.390                        & \textcolor{blue}{\underline{0.354}} & 0.385                        & \textcolor{blue}{\underline{0.259}} & 0.321                        & \textcolor{red}{\textbf{0.226}} & \textcolor{red}{\textbf{0.265}} & 0.164                        & 0.258                        & \textcolor{blue}{\underline{0.396}} & \textcolor{blue}{\underline{0.271}} & \textcolor{red}{\textbf{0.187}} & \textcolor{blue}{\underline{0.250}} & \textcolor{red}{\textbf{3}} (\textcolor{blue}{\underline{{6}}}) \\
&\texttt{iTransformer} (2023)           & 0.465                        & 0.470                        & 0.396                        & 0.422                        & 0.372                        & 0.401                        & 0.272                        & 0.332                        & 0.235                        & 0.274                        & \textcolor{blue}{\underline{0.161}} & 0.258                        & 0.398                        & 0.284                        & \textcolor{blue}{\underline{0.206}} & 0.269                        & \textcolor{red}{\textbf{0}} (\textcolor{blue}{\underline{{2}}}) \\ \midrule
&\method-t (Ours)         & \textcolor{red}{\textbf{0.417}} & \textcolor{blue}{\underline{0.430}} & 0.357                        & 0.399                        & 0.358                        & \textcolor{blue}{\underline{0.379}} & 0.259                        & \textcolor{blue}{\underline{0.316}} & \textcolor{blue}{\underline{0.231}} & \textcolor{blue}{\underline{0.269}} & 0.165                        & \textcolor{blue}{\underline{0.258}} & 0.419                        & 0.286                        & 0.232                        & 0.273                        & \textcolor{red}{\textbf{1}} (\textcolor{blue}{\underline{{6}}}) \\
&\method (Ours)       & \textcolor{blue}{\underline{0.418}} & \textcolor{red}{\textbf{0.421}} & 0.351                        & \textcolor{red}{\textbf{0.386}} & \textcolor{red}{\textbf{0.346}} & \textcolor{red}{\textbf{0.369}} & 0.265                        & 0.321                        & 0.240                        & 0.280                        & \textcolor{red}{\textbf{0.157}} & \textcolor{red}{\textbf{0.247}} & \textcolor{red}{\textbf{0.378}} & \textcolor{red}{\textbf{0.248}} & 0.213                        & \textcolor{red}{\textbf{0.245}} & \textcolor{red}{\textbf{9}} (\textcolor{blue}{\underline{{1}}})\\ \bottomrule[1pt]
\end{tabular}
}
\end{table*}

Table~\red{\ref{tab.light weight}} presents the LTSF performance of the 7 lightweight models, most of which are composed of linear layers. Table~\red{\ref{tab.complex model}} summarizes the LTSF performance of the 8 complex models, including MLP-based, CNN-based, and Transformer-based architectures. All models were trained 3 times with NVIDIA RTX 6000 Ada GPUs. The averaged MSE and MAE across all prediction horizons $H\in\{96,192,336,720\}$ are reported. The full results can be found in Appendix \red{\ref{app.FullResults}}.


From Table \red{\ref{tab.light weight}}, several key insights emerge: (1) From Fig. \red{\ref{fig.intro}}, the smallest models are \texttt{CMoS}, \texttt{FITS}, and \texttt{SparseTSF}, which usually encode a single hypothesis, {\em e.g.}, correlation among historical chunks in \texttt{CMoS}, simplifying their designs. The simplification may underfit some complex datasets, leading to their inferior performance in Table \ref{tab.light weight}; (2) The best baseline appears to be \texttt{PaiFilter}, whose parameter size is at a similar level as our \method, while being larger than \method-t, indicating a less effective use of parameters; (3) Our \method\ shows consistent superiority over the lightweight baselines, achieving 9 first-places (39 first-places in Table \red{\ref{tab.Full Lightweight model}} of Appendix \red{\ref{app.FullResults}}), confirming its non-trivial design inspired by LVM biases, and suggesting a best trade-off between performance and model size; and (4) Our \method-t further reduces the model size, thus is less powerful, but still achieves 6 second-places, surpassing the baselines in most cases.

Moreover, from Table \red{\ref{tab.complex model}}, complex models --- whose parameter sizes range from 400K to 10.5M (Fig. \red{\ref{fig.intro}}(c)), {\em i.e.}, $\sim 2\times$ to $ 50\times$ larger than \method\ --- do not show superiority over lightweight models. This observation is consistent with some existing studies \cite{lin2024sparsetsf,lin2024cyclenet}. The best complex baseline appears to be \texttt{PatchTST}, confirming its well-accepted design of time series patching (note: different from our patching, as discussed in $\S$\red{\ref{sec.ib2method}}) and channel-independence assumption. Whereas, it needs 3M+ parameters. In contrast, {\method}(-t) outperform these baselines in most cases,
particularly in handling datasets that contain anomalies such as ETTm1 (see $\S$\red{\ref{sec.exp.analysis}}).

These results underscore \method's potential as a powerful small model, while suggest \method-t as a less powerful yet more compact model for practical deployment in resource-constrained scenarios.


\subsection{Comparing with Pretrained Large Models}\label{sec.exp.large_models}

\begin{figure}[t!]
\centering
\includegraphics[width=0.75\columnwidth]{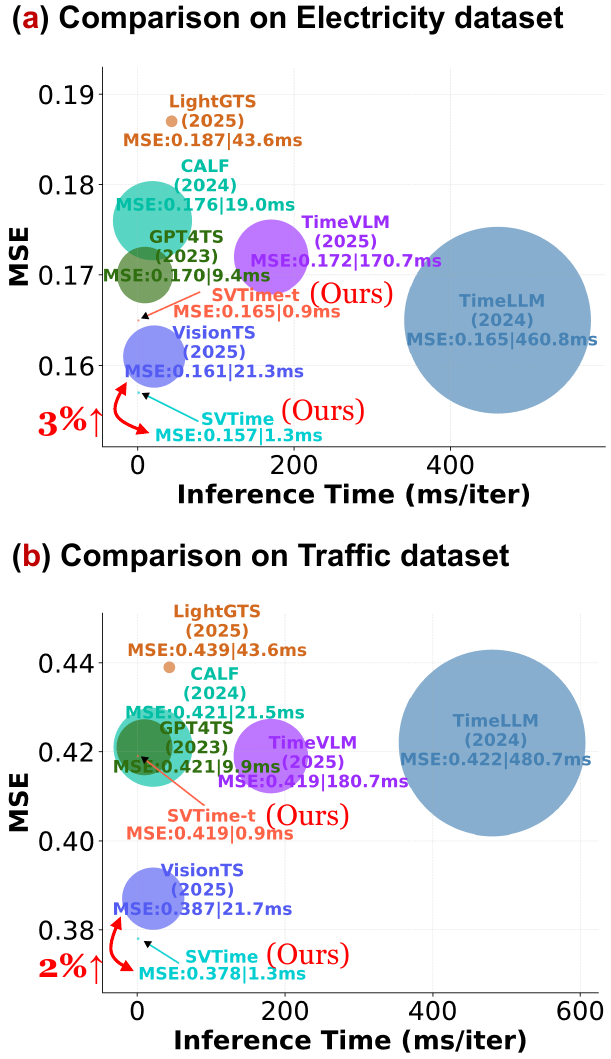}
\caption{LTSF performance comparison with pre-trained large models on (a) Electricity and (b) Traffic datasets. Bubble size is proportional to the parameter size of each model.}\label{fig.large_model}
\vspace{-0.2cm}
\end{figure}

Fig. \red{\ref{fig.large_model}}(a)(b) compare {\method}(-t) with the six pre-trained large models on Electricity and Traffic datasets after fine-tuning with the training sets. The full results on other datasets can be found in Appendix \red{\ref{app. Pre-trained Models}}. From the figures, we observe (1) Significant difference in parameter size and inference time between {\method}(-t) and other large models. For example, \method\ only uses $0.2\%$ parameters of \texttt{VisionTS}, and $0.003\%$ parameters of \texttt{TimeLLM}; (2) \texttt{VisionTS} performs better than other large models, confirming the rationale of using LVM to inspire our small models; (3) Competitive performance of \method, which shows $3\%$ ($2\%$) improvement over \texttt{VisionTS} on Electricity (Traffic) dataset, and may be the only small model in our experiments that can rival large models, without any pre-training; and (4) \method-t, despite its slightly lower performance than \texttt{VisionTS}, still outperforming some large models such as \texttt{GPT4TS}, \texttt{CALF}, and \texttt{LightGTS}, suggesting its potential in energy-saving scenarios.


\begin{table*}[htb]
\centering
\caption{Ablation analysis of {\method}(-t). MSE and MAE are averaged over different prediction lengths. Lower MSE and MAE are better. “Improvement”s of ablations (a)(b) (or (c)(d)(e)) are relative to \method\ (or \method-t).
}\label{tab. ablation}
\vspace{-0.1cm}
\resizebox{\textwidth}{!}{
\begin{tabular}{l|cc|cc|cc|cc|cc|cc|cc|cc}
\toprule[1pt]
\textbf{Method}           & \multicolumn{2}{c}{ETTh1}                                         & \multicolumn{2}{c}{ETTh2}                                       & \multicolumn{2}{c}{ETTm1}                                         & \multicolumn{2}{c}{ETTm2}                                         & \multicolumn{2}{c}{Weather}                                       & \multicolumn{2}{c}{Electricity}                                   & \multicolumn{2}{c}{Traffic}            & \multicolumn{2}{c}{Solar}                           \\
    & MSE                             & MAE                             & MSE                            & MAE                            & MSE                             & MAE                             & MSE                             & MAE                             & MSE                             & MAE                             & MSE                             & MAE                             & MSE                             & MAE        & MSE                             & MAE                       \\\midrule
\method            & 0.418                           & \textbf{0.421}                           & 0.351                          & \textbf{0.386}                          & \textbf{0.346}                           & \textbf{0.369 }                          & 0.265                           & 0.321                           & \textbf{0.240 }                          & 0.280                           & \textbf{0.157 }                          & \textbf{0.247}                           & \textbf{0.378}                           & \textbf{0.248}         &    \textbf{0.213 }                              &    \textbf{0.245}         \\\hline
(a) - IB2 & 0.420                           & 0.429                           & \textbf{0.342}                          & 0.389                          & 0.361                           & 0.380                           & \textbf{0.254 }                          & \textbf{0.314   }                        & 0.243                           & \textbf{0.279 }                          & 0.165                           & 0.257                           & 0.485                           & 0.344         &    0.233                               &    0.264        \\
Improvement        & \textcolor{red}{-0.48\%}  & \textcolor{red}{-1.90\%}  & 2.56\%                         & \textcolor{red}{-0.78\%} & \textcolor{red}{-4.34\%}  & \textcolor{red}{-2.98\%}  & 4.15\%                          & 2.18\%                          & \textcolor{red}{-1.25\%}  & 0.36\%                          & \textcolor{red}{-5.10\%}  & \textcolor{red}{-4.05\%}  & \textcolor{red}{-28.31\%} & \textcolor{red}{-38.71\%}                                 &             \textcolor{red}{-9.39\%}                           &     \textcolor{red}{-7.35\%}            \\\hline
(b) - Backcast  & \textbf{0.411}                           & \textbf{0.421}                           & 0.369                          & 0.402                          & 0.420                           & 0.428                           & 0.287                           & 0.340                           & 0.263                           & 0.301                           & 0.214                           & 0.295                           & 0.518                           & 0.282         &    0.418                               &    0.434        \\
Improvement        & 1.67\%                          & 0.00\%                          & \textcolor{red}{-5.13\%} & \textcolor{red}{-4.15\%} & \textcolor{red}{-21.39\%} & \textcolor{red}{-15.99\%} & \textcolor{red}{-8.30\%}  & \textcolor{red}{-5.92\%}  & \textcolor{red}{-9.58\%}  & \textcolor{red}{-7.50\%}  & \textcolor{red}{-36.31\%} & \textcolor{red}{-19.43\%} & \textcolor{red}{-37.04\%} & \textcolor{red}{-13.71\%}                                           &              \textcolor{red}{-96.24\%}                          &   \textcolor{red}{-77.14\%}              \\ \midrule
\method-t          &\textbf{0.417 }                          & 0.430                           &\textbf{0.357 }                         & \textbf{0.399 }                         & \textbf{0.358 }                          & 0.379                           & \textbf{0.259 }                          &\textbf{0.316 }                          & \textbf{0.231 }                          &\textbf{0.269 }                          & \textbf{0.165}                           & \textbf{0.258  }                         & \textbf{0.419 }                          & 0.286         &   \textbf{0.232 }                              &    0.273        \\\hline 
(c) -IB2 &0.422                           & \textbf{0.428 }                          & 0.366                          & 0.411                          & 0.361                           & \textbf{0.377}                           & 0.289                           & 0.336                           & 0.293                           & 0.315                           & 0.170                           & 0.261                           & 0.425                           & 0.284         &    0.251                               &    \textbf{0.272}        \\
Improvement        & \textcolor{red}{-1.20\%}  & 0.47\%                          & \textcolor{red}{-2.52\%} & \textcolor{red}{-3.01\%} & \textcolor{red}{-0.84\%}  & 0.53\%                          & \textcolor{red}{-11.58\%} & \textcolor{red}{-6.33\%}  & \textcolor{red}{-26.84\%} & \textcolor{red}{-17.10\%} & \textcolor{red}{-3.03\%}  & \textcolor{red}{-1.16\%}  & \textcolor{red}{-1.43\%}  & 0.70\%                                                               &                    \textcolor{red}{-8.19\%}                    &     0.37\%            \\\hline
(d) -IB3   & 0.428                           & 0.440                           & 0.359                          & \textbf{0.399 }                         & 0.362                           & 0.378                           & 0.289                           & 0.336                           & 0.243                           & 0.280                           & 0.169                           & 0.260                           & 0.425                           & \textbf{0.283}         &    0.251                               &    0.273        \\
Improvement        & \textcolor{red}{-2.64\%}  & \textcolor{red}{-2.33\%}  & \textcolor{red}{-0.56\%} & 0.00\%                         & \textcolor{red}{-1.12\%}  & 0.26\%                          & \textcolor{red}{-11.58\%} & \textcolor{red}{-6.33\%}  & \textcolor{red}{-5.19\%}  & \textcolor{red}{-4.09\%}  & \textcolor{red}{-2.42\%}  & \textcolor{red}{-0.78\%}  & \textcolor{red}{-1.43\%}  & 1.05\%                                                               &                     \textcolor{red}{-8.19\%}                    &    0.00\%             \\\hline
(e) - Backcast  & 0.488                           & 0.510                           & 0.371                          & 0.404                          & 0.565                           & 0.449                           & 0.313                           & 0.354                           & 0.295                           & 0.319                           & 0.212                           & 0.293                           & 0.622                           & 0.387         &    0.385                               &    0.423        \\
Improvement        & \textcolor{red}{-17.03\%} & \textcolor{red}{-18.60\%} & \textcolor{red}{-3.92\%} & \textcolor{red}{-1.25\%} & \textcolor{red}{-57.82\%} & \textcolor{red}{-18.47\%} & \textcolor{red}{-20.85\%} & \textcolor{red}{-12.03\%} & \textcolor{red}{-27.71\%} & \textcolor{red}{-18.59\%} & \textcolor{red}{-28.48\%} & \textcolor{red}{-13.57\%} & \textcolor{red}{-48.45\%} & \textcolor{red}{-35.31\%}                                                     &                    \textcolor{red}{-65.95\%}                    &         \textcolor{red}{-54.95\%}          \\
\bottomrule[1pt]
\end{tabular}
}
\vspace{-0.1cm}
\end{table*}

\subsection{Ablation Analysis}\label{sec.exp.ablation}

We validate the design of {\method}(-t) through ablation studies on all datasets. Table \ref{tab. ablation} summarizes the analysis: (a) removes IB2 from \method, {\em i.e.}, without using $\mat{W}_{1}$, ..., $\mat{W}_{K}$ for enabling patch-wise variety. This is equivalent to the model with only IB1 ($\S$\red{\ref{sec.ib1method}}), {\em i.e.}, using a single $\mat{W}$; (b) removes the backcast-residual decomposition framework ($\S$\red{\ref{sec.backcast}}) from \method; (c) removes IB2 from \method-t, which is equivalent to set $K=1$ for $\mat{\tilde{W}}_{1}$, ..., $\mat{\tilde{W}}_{K}$ (see $\S$\red{\ref{sec.ib3method}}), {\em i.e.}, disabling patch-wise variety; (d) removes IB3 from \method-t, {\em e.g.}, removing the annealing constraint function by simply setting $w(j,n)=1$ in Eq. (\red{\ref{eq.constraint}}) for $\forall 1\le j\le M$, $1\le n\le N$, which is equivalent to set $\mat{\tilde{W}}_{k}=w_{k}^{p}$ for $1\le k\le K$; and (e) removes the backcast-residual decomposition framework from \method-t. Note that IB1 cannot be removed as it establishes the basis of the proposed models.

Table \red{\ref{tab. ablation}} reveals several key insights into the design of {\method}(-t). In (a)(c), removing IB2 --- patch-wise variety --- from the modeling of historical periods degrades the performance of {\method}(-t) in most cases, confirming the effectiveness of fine-grained modeling of within-period patches using $\mat{W}_{k}$ (or $\mat{\tilde{W}_{k}}$) ($1\le k \le K$). In (b)(e), we observe a major performance drop by removing the backcast-residual decomposition, highlighting the disadvantage of solely modeling periodical patterns while suggesting the effectiveness of the adaptive decomposition in mitigating this bias. Finally, (d) underscores the importance of the proposed annealing constraint function: removing it leads to uniform attention to historical periods, which contradicts LVM's distance-attenuating local attention, indirectly validating IB3's usefulness in LVM forecasters.

Overall, the patch-wise variety, distance-attenuating local attention, and the backcast-residual decomposition are crucial to the success of \method\ and \method-t.

\subsection{Comparing with Knowledge Distillation}\label{sec.exp.kd}

\begin{table}[t!]
\centering
\caption{Comparison between {\method}(-t) and \texttt{OccamVTS}.}\label{tab. distilation}
\resizebox{0.9\columnwidth}{!}{
\begin{tabular}{l|rrr}
\toprule[1pt]
\textbf{Metrics}                 & \method & \method-t & \texttt{OccamVTS}  \\\midrule
MSE                    & \red{\textbf{0.346}}                  & 0.358                    & \textcolor{blue}{\underline{0.349}}    \\
MAE                    & \red{\textbf{0.369}}                  & 0.379                    & \textcolor{blue}{\underline{0.372}}    \\\hline
GPU memory (MiB)          & 1,095                   & 692                      & 11,859     \\
Train time (s/epoch)   & 2.9                    & 2.6                      & 275       \\
Inference time (ms/iter) & 1.08                   & 0.84                     & 2.83      \\
Parameter size       & 215.5K                 & 162.7K                   & 2,834.4K\\
\bottomrule[1pt]
\end{tabular}
}
\vspace{-0.1cm}
\end{table}

\begin{figure*}[t!]
\centering
\includegraphics[width=0.9\textwidth]{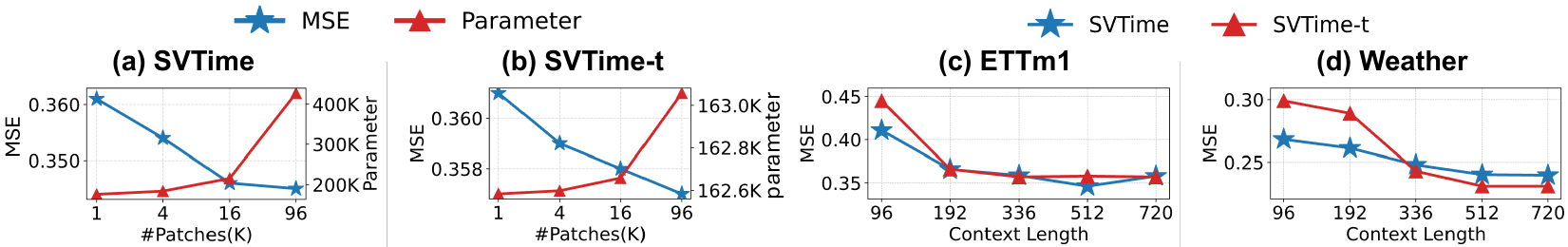}
\caption{Results on the impact of the number of patches $K$ on (a) \method; and (b) \method-t in terms of LTSF performance and model size; and the impact of lookback window (or context) length on both models using (c) ETTm1; and (d) Weather datasets.}\label{fig.patch and windows}
\vspace{-0.1cm}
\end{figure*}

\begin{figure*}[t!]
\centering
\includegraphics[width=0.85\textwidth]{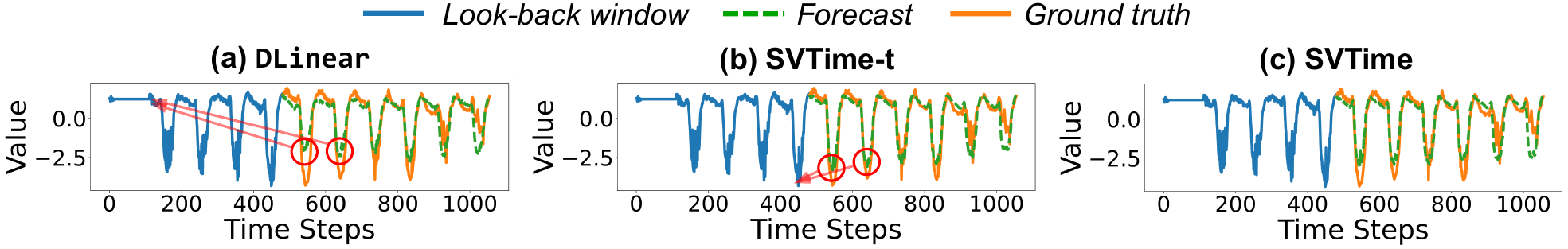}
\caption{Comparison of (a) \texttt{DLinear}; (b) \method-t; and (c) \method\ on the same example in the ETTm1 dataset that highlights the importance of distance-attenuating local attention.}\label{fig.ib3 example}
\end{figure*}

We compare {\method}(-t) with \texttt{OccamVTS} \cite{lyu2025occamvts} --- a knowledge distillation (KD) based student model trained using \texttt{MAE} as its teacher model --- using EETm1 dataset as an example (results on other datasets are similar thus are omitted for brevity).

Table \red{\ref{tab. distilation}} summarizes the comparison in terms of both LTSF performance and computational costs. All models were trained with the same batch size of 512. The results are averaged over all prediction horizons $H \in \{96,192,336,720\}$. From Table \red{\ref{tab. distilation}}, we observe (1) all of the three models demonstrate comparable performance, while \method\ slightly outperforms \texttt{OccamVTS}, suggesting all of the models inherit the merits of the LVM forecaster, either through ``physics'' guided learning or KD; (2) \texttt{OccamVTS} may not be considered as a lightweight model for its 2.8M parameter size. This is because it uses Transformer backbone --- the same as its teacher model --- and doesn't support cross-architecture KD; (3) \method-t is the most efficient model with $17\times$ less memory, $106\times$ faster training, and $3\times$ faster inference than \texttt{OccamVTS}. Also, \method's cost is more appealing than \texttt{OccamVTS}. The high training cost of \texttt{OccamVTS} is due to its involvement of a large teacher model \texttt{MAE}; and (4) we note that \texttt{OccamVTS}'s reliance on a large teacher model needs extra costs for training and fine-tuning the teacher model before KD. As such, {\method}(-t) offers advantages over KD-based models.

\subsection{Performance Analysis}\label{sec.exp.analysis}

In this section, we perform an in-depth analysis of {\method}(-t).



\vspace{0.2cm}

\noindent{\textbf{Impact of the number of patches}}. Fig.\red{\ref{fig.patch and windows}}(a)(b) assess the impact of the number of within-period patches, {\em i.e.}, $K$ ($\S$\red{\ref{sec.ib2method}}) on the performance of \method\ and \method-t when fixing period length $P=96$ on the ETTm1 dataset. Also, since increasing $K$ leads to more parameters in $\mat{W}_{1}$, ..., $\mat{W}_{K}$ (for \method) and $\mat{\tilde{W}}_{1}$, ..., $\mat{\tilde{W}}_{K}$ (for \method-t), we evaluate their parameter sizes. From Fig.\red{\ref{fig.patch and windows}}(a)(b), both models achieve better performance with a larger $K$, which corresponds to more fine-grained patching. Notably, for \method, the performance improvement from $K=16$ to $K=96$ is small while the model size expands a lot, suggesting a saturating point at $K=16$. For \method-t, the changing ranges of MSE and model size caused by varying $K$ are relatively small. This is because $\mat{\tilde{W}}_{1}$, ..., $\mat{\tilde{W}}_{K}$ only have $O(K)$ learnable parameters, thus varying $K$ has a small impact. Overall, for both models, we set $K=\lfloor P/6\rfloor$ for the best trade-off between performance and efficiency.



\vspace{0.2cm}

\noindent{\textbf{Distance-attenuating local attention}}. Fig. \red{\ref{fig.ib3 example}} compares \method\, \method-t, and \texttt{DLinear} on a sample ETTm1 case that highlights the importance of distance-attenuating local attention ($\S$\red{\ref{sec.ib3}}). The lookback window starts with a flattened high value and ends with a notable dip. \texttt{DLinear} uses a linear layer to forecast without any constraint, thus doesn't guarantee local attention. In fact, Fig. \red{\ref{fig.ib3 example}}(a) shows its first two predicted dips might be influenced by the high value in the lookback window, while the first two ground truth dips follow the nearby dip in the lookback window, leading to errors. In contrast, \method-t effectively uses its constraint function to predict the first two dips. \method\ slightly underperforms but is still better than \texttt{DLinear}. Moreover, \method-t's forecast of all dips gradually rises due to its attenuating local attention, which matches the ground truth. However, \texttt{DLinear} has no such pattern.

\vspace{0.2cm}


\noindent{\textbf{Impact of lookback window}}. Fig. \red{\ref{fig.patch and windows}}(c)(d) assess {\method}(-t)'s performance {\em w.r.t.} varying lookback window length on ETTm1 and Weather datasets. Using MSE metric, we observe both \method\ and \method-t benefit from longer lookback windows. \method\ is more effective than \method-t when the lookback window is small. This is possibly because a shorter lookback window may lead to more constrained local weights by \method-t's annealing constraint function, resulting in a slight sensitivity to a small context length.

\section{Conclusion}

This paper introduces \method\ and \method-t, novel lightweight models whose design is informed by the inductive biases --- analogous to the ``physics'' governing the behaviors --- of a powerful LVM forecaster. Through the discovery and re-programming of three inductive biases, {\em i.e.}, inter-period consistency, patch-wise variety, and distance-attenuating local attention, along with the adoption of a backcast-residual decomposition framework, {\method}(-t) effectively inherits the merits of the LVM forecaster, while mitigating the dominant bias of period-prone forecasting. Extensive experiments on 8 benchmark datasets with 21 SOTA baselines covering lightweight, complex, pre-trained large models, and a KD model demonstrate {\method}(-t)'s large-model-like performance and economical size, highlighting the potential in energy-saving scenarios.


\clearpage
\bibliographystyle{ACM-Reference-Format}
\bibliography{refer}

\appendix
\clearpage
\section{Experiment Setting}

\subsection{Datasets}\label{app.Datasets}
Following \cite{liu2023itransformer,wu2023timesnet,liu2022scinet,si2025cmos}, our experiments are conducted on 8 widely used LTSF benchmark datasets that cover a wide range of sampling frequencies, number of variates, levels of periodicity, and real-world domains. The four ETT datasets (ETTh1, ETTh2, ETTm1, ETTm2) record oil temperature from two electric transformers, sampled at 15-minute and hourly intervals. The Weather dataset collects measurements of meteorological indicators in Germany every 10 minutes. The Electricity dataset records hourly electricity consumption of Portuguese clients. The Traffic dataset measures hourly road occupancy rates from sensors on San Francisco freeways. The Solar contains hourly solar power output measurements from U.S. photovoltaic plants. Table~\ref{tab. dataset details} summarizes the statistics of the datasets.

\begin{table}[htbp]
\centering
\caption{Statistics of the benchmark datasets. ``Dataset Size'' is organized in (Train, Validation, Test).}
\label{tab. dataset details}
\resizebox{\columnwidth}{!}{
\begin{tabular}{lcccc}
\toprule[1pt]
\textbf{Dataset} & \textbf{\# Variates} & \textbf{Series Length} & \textbf{Dataset Size}          & \textbf{Frequency} \\ \midrule
ETTh1       & 7       & 17420         & (8545, 2881, 2881)    & Hourly      \\
ETTh2       & 7       & 17420         & (8545, 2881, 2881)    & Hourly      \\
ETTm1       & 7       & 69680         & (34465, 11521, 11521) & 15 mins     \\
ETTm2       & 7       & 69680         & (34465, 11521, 11521) & 15 mins     \\
Weather     & 321     & 52696         & (36792, 5271, 10540)  & 10 mins     \\
Electricity & 21      & 26304         & (18317, 2633, 5261)   & Hourly     \\ 
Traffic     & 862     & 17544         & (12185, 1757, 3509)   & Hourly      \\

Solar     &   137   &   52560       & (36792,5768,11024)   & Hourly      \\
\bottomrule[1pt]
\end{tabular}}
\end{table}

\subsection{Running Environment}
The experiments are conducted on a Linux server (kernel 5.15.0-139) with NVIDIA RTX 6000 Ada GPUs (48 GB). The environment uses Python 3.12.8, PyTorch 2.5.1 with CUDA 12.4 and cuDNN 9.1. The key libraries include NumPy 2.1.3, Pandas 2.2.3, Matplotlib 3.10.0, SciPy 1.15.1, scikit-learn 1.6.1, and torchvision 0.20.1.

\section{More Experimental Results}
\subsection{Statistical Stability}\label{app.Statistical Stability}
To ensure the statistical reliability of our results, we trained and evaluated all models three times on each benchmark dataset using different random seeds ($seed=2021\sim2023$). Table~\red{\ref{tab. statistical Stability}} reports the average standard deviations of \method(-t) across the eight datasets. As shown in the table, the maximum standard deviation is below 0.004, indicating that SVTime(-t) exhibits strong stability and robustness.

\begin{table}[htb]
\caption{Standard deviations of \method(-t) across all baselines. Results computed over three independent runs with random seeds ($seed=2021\sim2023$).}
\label{tab. statistical Stability}
\resizebox{\columnwidth}{!}{
\begin{tabular}{c|cc|cc}
\toprule[1pt]
                            & \multicolumn{2}{c}{\method-t}&          \multicolumn{2}{c}{\method}                   \\
Dataset                     & MSE                 & MAE               & MSE                & MAE               \\ \midrule
ETTh1                       & 0.417   $\pm$ 0.001 & 0.430 $\pm$ 0.001 & 0.418  $\pm$ 0.001 & 0.421 $\pm$ 0.001 \\
ETTh2                       & 0.358   $\pm$ 0.003 & 0.399 $\pm$ 0.001 & 0.351  $\pm$ 0.003 & 0.386 $\pm$ 0.001 \\
ETTm1                       & 0.358   $\pm$ 0.002 & 0.379 $\pm$ 0.001 & 0.346  $\pm$ 0.002 & 0.369 $\pm$ 0.001 \\
ETTm2                       & 0.260   $\pm$ 0.001 & 0.316 $\pm$ 0.001 & 0.265  $\pm$ 0.004 & 0.321 $\pm$ 0.002 \\
Weather                     & 0.231   $\pm$ 0.001 & 0.269 $\pm$ 0.001 & 0.240  $\pm$ 0.002 & 0.280 $\pm$ 0.003 \\
Electricity                 & 0.165   $\pm$ 0.000 & 0.258 $\pm$ 0.000 & 0.157  $\pm$ 0.003 & 0.247 $\pm$ 0.003 \\
Traffic                     & 0.419   $\pm$ 0.000 & 0.287 $\pm$ 0.001 & 0.378  $\pm$ 0.001 & 0.248 $\pm$ 0.003 \\
Solar                       & 0.232   $\pm$ 0.001 & 0.274 $\pm$ 0.001 & 0.213  $\pm$ 0.001 & 0.245 $\pm$ 0.002 \\ \bottomrule[1pt]
\end{tabular}
}
\end{table}

\subsection{Full Results with baseline Models}\label{app.FullResults}
Tables~\red{\ref{tab.Full Lightweight model}} and Table~\red{\ref{tab. Full Complex Models}} provide the complete comparison of \method(-t) with seven light-weight models and eight complex models from different categories, respectively, serving as a supplement to Table~\red{\ref{tab.light weight}} and Table~\red{\ref{tab.complex model}} in the main paper.

From these tables, we can observe that \method maintains a clear advantage when compared against all baseline methods.
Against the light-weight models, it achieves 39 first-place results; against the complex models, it achieves 35 first-place results.
Although \method-t shows some performance degradation, it still obtains 19 second-best or better results when compared with the light-weight models and 26 second-best or better results when compared with the complex models.
These results demonstrate that our \method(-t) as a light-weight model, remains highly competitive with both light-weight and complex baselines.
And these results further validate that incorporating inductive biases from vision models is an effective strategy for designing small yet powerful models.

\begin{table*}[p]
\centering
\caption{Full LTSF results on the benchmark datasets with Light-weight Models. Lower MSE and MAE indicate better forecasting accuracy. \textcolor{red}{\textbf{Red}} denote the best performance,  \textcolor{blue}{\underline{blue}} indicate the second-best results.}\label{tab.Full Lightweight model}
\resizebox{0.75\textwidth}{!}{
\begin{tabular}{c|c|cc|cc|cc|cc|cc|cc|cc|cc|cc}
\toprule[1pt]
                              &           & \multicolumn{2}{c}{\method-t}                                & \multicolumn{2}{c}{\method}                                   & \multicolumn{2}{|c}{\texttt{Dlinear}}                            & \multicolumn{2}{c}{\texttt{FITS}}                                    & \multicolumn{2}{c}{\texttt{TexFilter}}                               & \multicolumn{2}{c}{\texttt{PaiFilter}}                               & \multicolumn{2}{c}{\texttt{SparceTSF}}                               & \multicolumn{2}{c}{\texttt{CycleNet}}                                & \multicolumn{2}{c}{\texttt{CMoS}}                                    \\
                       & Pred\_len & \multicolumn{1}{c}{MSE}      & \multicolumn{1}{c}{MAE}      & \multicolumn{1}{c}{MSE}      & \multicolumn{1}{c}{MAE}      & \multicolumn{1}{|c}{MSE}      & \multicolumn{1}{c}{MAE} & \multicolumn{1}{c}{MSE}      & \multicolumn{1}{c}{MAE}      & \multicolumn{1}{c}{MSE}      & \multicolumn{1}{c}{MAE}      & \multicolumn{1}{c}{MSE}      & \multicolumn{1}{c}{MAE}      & \multicolumn{1}{c}{MSE}      & \multicolumn{1}{c}{MAE}      & \multicolumn{1}{c}{MSE}      & \multicolumn{1}{c}{MAE}      & \multicolumn{1}{c}{MSE}      & \multicolumn{1}{c}{MAE}      \\\midrule
                              & 96        & \textcolor{blue}{\underline{0.368}} & \textcolor{blue}{\underline{0.394}} & \textcolor{red}{\textbf{0.361}} & \textcolor{red}{\textbf{0.387}} & 0.379                        & 0.405                   & 0.396                        & 0.414                        & 0.390                        & 0.422                        & 0.385                        & 0.411                        & 0.408                        & 0.425                        & 0.374                        & 0.398                        & 0.377                        & 0.401                        \\
                              & 192       & \textcolor{blue}{\underline{0.406}} & 0.418                        & 0.409                        & \textcolor{red}{\textbf{0.408}} & 0.448                        & 0.461                   & 0.430                        & 0.435                        & 0.418                        & 0.437                        & 0.421                        & 0.433                        & 0.438                        & 0.447                        & \textcolor{red}{\textbf{0.404}} & \textcolor{blue}{\underline{0.417}} & 0.410                        & 0.422                        \\
                              & 336       & 0.431                        & \textcolor{blue}{\underline{0.434}} & 0.445                        & \textcolor{red}{\textbf{0.428}} & 0.449                        & 0.452                   & 0.446                        & 0.444                        & 0.432                        & 0.449                        & 0.437                        & 0.445                        & \textcolor{red}{\textbf{0.423}} & 0.444                        & 0.435                        & 0.437                        & \textcolor{blue}{\underline{0.426}} & 0.434                        \\
\multirow{-4}{*}{\rotatebox{90}{ETTh1}}       & 720       & 0.464                        & 0.472                        & 0.457                        & \textcolor{red}{\textbf{0.461}} & 0.512                        & 0.519                   & \textcolor{blue}{\underline{0.448}} & 0.462                        & 0.457                        & 0.477                        & 0.474                        & 0.481                        & \textcolor{red}{\textbf{0.433}} & \textcolor{blue}{\underline{0.461}} & 0.464                        & 0.472                        & 0.451                        & 0.466                        \\\midrule
                              & 96        & 0.283                        & 0.344                        & \textcolor{blue}{\underline{0.283}} & \textcolor{red}{\textbf{0.329}} & \textcolor{red}{\textbf{0.280}} & 0.347                   & 0.286                        & 0.349                        & 0.303                        & 0.364                        & 0.303                        & 0.361                        & 0.311                        & 0.362                        & 0.308                        & \textcolor{blue}{\underline{0.339}} & 0.306                        & 0.363                        \\
                              & 192       & 0.349                        & 0.388                        & 0.348                        & \textcolor{red}{\textbf{0.374}} & 0.355                        & 0.399                   & \textcolor{blue}{\underline{0.345}} & 0.386                        & 0.370                        & 0.402                        & 0.366                        & 0.399                        & 0.364                        & 0.397                        & \textcolor{red}{\textbf{0.335}} & \textcolor{blue}{\underline{0.382}} & 0.355                        & 0.395                        \\
                              & 336       & 0.379                        & 0.416                        & \textcolor{red}{\textbf{0.364}} & \textcolor{red}{\textbf{0.395}} & 0.435                        & 0.453                   & \textcolor{blue}{\underline{0.364}} & \textcolor{blue}{\underline{0.403}} & 0.391                        & 0.424                        & 0.389                        & 0.426                        & 0.373                        & 0.408                        & 0.369                        & 0.413                        & 0.367                        & 0.407                        \\
\multirow{-4}{*}{\rotatebox{90}{ETTh2}}       & 720       & 0.419                        & 0.448                        & 0.408                        & 0.446                        & 0.697                        & 0.591                   & \textcolor{red}{\textbf{0.389}} & \textcolor{red}{\textbf{0.429}} & 0.423                        & 0.449                        & 0.425                        & 0.452                        & 0.393                        & \textcolor{blue}{\underline{0.430}} & 0.419                        & 0.449                        & \textcolor{blue}{\underline{0.390}} & 0.432                        \\\midrule
                              & 96        & \textcolor{red}{\textbf{0.303}} & \textcolor{blue}{\underline{0.348}} & 0.305                        & \textcolor{red}{\textbf{0.341}} & 0.312                        & 0.359                   & 0.318                        & 0.358                        & 0.313                        & 0.366                        & 0.305                        & 0.359                        & 0.313                        & 0.353                        & 0.309                        & 0.354                        & \textcolor{blue}{\underline{0.305}} & 0.352                        \\
                              & 192       & \textcolor{blue}{\underline{0.337}} & \textcolor{blue}{\underline{0.368}} & \textcolor{red}{\textbf{0.320}} & \textcolor{red}{\textbf{0.351}} & 0.350                        & 0.383                   & 0.348                        & 0.375                        & 0.352                        & 0.389                        & 0.340                        & 0.380                        & 0.342                        & 0.370                        & 0.350                        & 0.377                        & 0.341                        & 0.371                        \\
                              & 336       & \textcolor{blue}{\underline{0.369}} & \textcolor{blue}{\underline{0.385}} & \textcolor{red}{\textbf{0.353}} & \textcolor{red}{\textbf{0.378}} & 0.378                        & 0.400                   & 0.379                        & 0.392                        & 0.380                        & 0.404                        & 0.370                        & 0.395                        & 0.373                        & 0.388                        & 0.372                        & 0.391                        & 0.372                        & 0.388                        \\
\multirow{-4}{*}{\rotatebox{90}{ETTm1}}       & 720       & \textcolor{blue}{\underline{0.422}} & \textcolor{blue}{\underline{0.414}} & \textcolor{red}{\textbf{0.407}} & \textcolor{red}{\textbf{0.405}} & 0.439                        & 0.441                   & 0.431                        & 0.420                        & 0.439                        & 0.438                        & 0.439                        & 0.427                        & 0.423                        & 0.415                        & 0.426                        & 0.421                        & 0.433                        & 0.423                        \\\midrule
                              & 96        & \textcolor{red}{\textbf{0.165}} & \textcolor{blue}{\underline{0.254}} & 0.182                        & \textcolor{red}{\textbf{0.253}} & 0.168                        & 0.261                   & 0.169                        & 0.259                        & 0.188                        & 0.276                        & 0.179                        & 0.265                        & 0.166                        & 0.256                        & \textcolor{blue}{\underline{0.166}} & 0.254                        & 0.177                        & 0.264                        \\
                              & 192       & 0.224                        & \textcolor{blue}{\underline{0.293}} & 0.243                        & 0.304                        & 0.227                        & 0.306                   & 0.223                        & 0.295                        & 0.250                        & 0.317                        & 0.235                        & 0.308                        & \textcolor{blue}{\underline{0.223}} & \textcolor{blue}{\underline{0.293}} & \textcolor{red}{\textbf{0.220}} & \textcolor{red}{\textbf{0.292}} & 0.233                        & 0.302                        \\
                              & 336       & 0.280                        & 0.331                        & 0.294                        & 0.341                        & 0.281                        & 0.342                   & \textcolor{blue}{\underline{0.275}} & \textcolor{blue}{\underline{0.329}} & 0.314                        & 0.358                        & 0.281                        & 0.338                        & \textcolor{red}{\textbf{0.273}} & \textcolor{red}{\textbf{0.326}} & 0.281                        & 0.334                        & 0.286                        & 0.336                        \\
\multirow{-4}{*}{\rotatebox{90}{ETTm2}}       & 720       & 0.369                        & 0.387                        & \textcolor{red}{\textbf{0.339}} & 0.387                        & 0.381                        & 0.407                   & 0.363                        & \textcolor{blue}{\underline{0.383}} & 0.400                        & 0.409                        & 0.364                        & 0.392                        & \textcolor{blue}{\underline{0.362}} & \textcolor{red}{\textbf{0.382}} & 0.366                        & 0.388                        & 0.371                        & 0.389                        \\\midrule
                              & 96        & 0.156                        & \textcolor{blue}{\underline{0.207}} & 0.169                        & 0.223                        & 0.173                        & 0.236                   & 0.174                        & 0.228                        & \textcolor{blue}{\underline{0.154}} & 0.208                        & \textcolor{red}{\textbf{0.145}} & \textcolor{red}{\textbf{0.198}} & 0.173                        & 0.227                        & 0.170                        & 0.224                        & 0.158                        & 0.209                        \\
                              & 192       & \textcolor{blue}{\underline{0.198}} & \textcolor{blue}{\underline{0.245}} & 0.214                        & 0.262                        & 0.220                        & 0.282                   & 0.216                        & 0.263                        & 0.199                        & 0.248                        & \textcolor{red}{\textbf{0.190}} & \textcolor{red}{\textbf{0.240}} & 0.215                        & 0.261                        & 0.212                        & 0.260                        & 0.199                        & 0.247                        \\
                              & 336       & 0.250                        & 0.286                        & 0.257                        & 0.293                        & 0.262                        & 0.316                   & 0.262                        & 0.296                        & 0.249                        & \textcolor{blue}{\underline{0.285}} & \textcolor{red}{\textbf{0.241}} & \textcolor{red}{\textbf{0.278}} & 0.261                        & 0.295                        & 0.258                        & 0.293                        & \textcolor{blue}{\underline{0.248}} & \textcolor{blue}{\underline{0.285}} \\
\multirow{-4}{*}{\rotatebox{90}{Weather}}     & 720       & 0.321                        & 0.336                        & 0.321                        & 0.340                        & 0.325                        & 0.367                   & 0.327                        & 0.343                        & 0.328                        & 0.339                        & \textcolor{red}{\textbf{0.316}} & \textcolor{red}{\textbf{0.332}} & 0.327                        & 0.343                        & 0.323                        & 0.339                        & \textcolor{blue}{\underline{0.319}} & \textcolor{blue}{\underline{0.334}} \\\midrule
                              & 96        & 0.135                        & 0.231                        & \textcolor{red}{\textbf{0.127}} & \textcolor{red}{\textbf{0.221}} & 0.136                        & 0.234                   & 0.273                        & 0.385                        & 0.132                        & 0.231                        & 0.132                        & 0.230                        & 0.168                        & 0.266                        & \textcolor{blue}{\underline{0.128}} & \textcolor{blue}{\underline{0.223}} & 0.137                        & 0.231                        \\
                              & 192       & 0.151                        & 0.245                        & \textcolor{red}{\textbf{0.143}} & \textcolor{red}{\textbf{0.234}} & 0.151                        & 0.248                   & 0.284                        & 0.392                        & 0.155                        & 0.251                        & 0.150                        & 0.246                        & 0.188                        & 0.285                        & \textcolor{blue}{\underline{0.144}} & \textcolor{blue}{\underline{0.237}} & 0.153                        & 0.246                        \\
                              & 336       & 0.167                        & 0.261                        & \textcolor{red}{\textbf{0.158}} & \textcolor{red}{\textbf{0.248}} & 0.166                        & 0.266                   & 0.298                        & 0.401                        & 0.171                        & 0.270                        & 0.167                        & 0.263                        & 0.201                        & 0.298                        & \textcolor{blue}{\underline{0.160}} & \textcolor{blue}{\underline{0.256}} & 0.166                        & 0.259                        \\
\multirow{-4}{*}{\rotatebox{90}{Electricity}} & 720       & 0.207                        & 0.295                        & \textcolor{red}{\textbf{0.198}} & \textcolor{red}{\textbf{0.283}} & 0.200                        & 0.298                   & 0.331                        & 0.420                        & 0.209                        & 0.304                        & 0.209                        & 0.299                        & 0.229                        & 0.319                        & \textcolor{blue}{\underline{0.198}} & \textcolor{blue}{\underline{0.288}} & 0.206                        & 0.293                        \\\midrule
                              & 96        & 0.395                        & 0.275                        & \textcolor{red}{\textbf{0.353}} & \textcolor{red}{\textbf{0.231}} & 0.398                        & 0.282                   & 0.411                        & 0.298                        & \textcolor{blue}{\underline{0.379}} & 0.280                        & 0.385                        & 0.277                        & 0.411                        & 0.285                        & 0.389                        & 0.275                        & 0.401                        & \textcolor{blue}{\underline{0.274}} \\
                              & 192       & 0.408                        & 0.281                        & \textcolor{red}{\textbf{0.364}} & \textcolor{red}{\textbf{0.243}} & 0.410                        & 0.287                   & 0.420                        & 0.298                        & 0.407                        & 0.296                        & 0.406                        & 0.288                        & 0.425                        & 0.295                        & \textcolor{blue}{\underline{0.403}} & 0.281                        & 0.414                        & \textcolor{blue}{\underline{0.279}} \\
                              & 336       & 0.418                        & 0.285                        & \textcolor{red}{\textbf{0.375}} & \textcolor{red}{\textbf{0.248}} & 0.420                        & 0.293                   & 0.429                        & 0.302                        & 0.419                        & 0.302                        & 0.417                        & 0.294                        & 0.429                        & 0.296                        & \textcolor{blue}{\underline{0.410}} & \textcolor{blue}{\underline{0.285}} & 0.424                        & 0.286                        \\
\multirow{-4}{*}{\rotatebox{90}{Traffic}}     & 720       & 0.454                        & \textcolor{blue}{\underline{0.305}} & \textcolor{red}{\textbf{0.421}} & \textcolor{red}{\textbf{0.271}} & 0.457                        & 0.314                   & 0.466                        & 0.322                        & 0.453                        & 0.318                        & 0.457                        & 0.317                        & 0.465                        & 0.315                        & \textcolor{blue}{\underline{0.446}} & 0.305                        & 0.460                        & 0.305                        \\\midrule
                              & 96        & 0.206                        & 0.257                        & \textcolor{red}{\textbf{0.183}} & \textcolor{red}{\textbf{0.238}} & 0.206                        & 0.279                   & 0.778                        & 0.706                        & 0.189                        & 0.253                        & \textcolor{blue}{\underline{0.184}} & \textcolor{blue}{\underline{0.240}} & 0.213                        & 0.257                        & 0.204                        & 0.273                        & 0.204                        & 0.252                        \\
                              & 192       & 0.229                        & 0.271                        & \textcolor{blue}{\underline{0.213}} & \textcolor{red}{\textbf{0.243}} & 0.227                        & 0.293                   & 0.777                        & 0.705                        & 0.215                        & 0.280                        & \textcolor{red}{\textbf{0.199}} & \textcolor{blue}{\underline{0.255}} & 0.233                        & 0.266                        & 0.225                        & 0.289                        & 0.222                        & 0.259                        \\
                              & 336       & 0.242                        & 0.281                        & 0.228                        & \textcolor{red}{\textbf{0.243}} & 0.242                        & 0.304                   & 0.775                        & 0.705                        & \textcolor{blue}{\underline{0.210}} & 0.271                        & \textcolor{red}{\textbf{0.204}} & \textcolor{blue}{\underline{0.260}} & 0.248                        & 0.275                        & 0.239                        & 0.293                        & 0.232                        & 0.269                        \\
\multirow{-4}{*}{\rotatebox{90}{Solar}}       & 720       & 0.250                        & 0.285                        & 0.228                        & \textcolor{red}{\textbf{0.254}} & 0.251                        & 0.311                   & 0.766                        & 0.703                        & \textcolor{blue}{\underline{0.220}} & 0.278                        & \textcolor{red}{\textbf{0.209}} & \textcolor{blue}{\underline{0.264}} & 0.254                        & 0.276                        & 0.248                        & 0.301                        & 0.237                        & 0.270                        \\\midrule
\multicolumn{2}{c}{Wins}                  & \multicolumn{2}{c}{\textcolor{red}{\textbf{2}} (\textcolor{blue}{\underline{17}})}                                   & \multicolumn{2}{c}{\textcolor{red}{\textbf{39}} (\textcolor{blue}{\underline{2}})}                                   & \multicolumn{2}{c}{\textcolor{red}{\textbf{1}} (\textcolor{blue}{\underline{0}})}                               & \multicolumn{2}{c}{\textcolor{red}{\textbf{2}} (\textcolor{blue}{\underline{7}})}                                    & \multicolumn{2}{c}{\textcolor{red}{\textbf{0}} (\textcolor{blue}{\underline{5}})}                                    & \multicolumn{2}{c}{\textcolor{red}{\textbf{11}} (\textcolor{blue}{\underline{5}})}                                   & \multicolumn{2}{c}{\textcolor{red}{\textbf{5}} (\textcolor{blue}{\underline{5}})}                                    & \multicolumn{2}{c}{\textcolor{red}{\textbf{4}} (\textcolor{blue}{\underline{16}})}                                   & \multicolumn{2}{c}{\textcolor{red}{\textbf{0}}    (\textcolor{blue}{\underline{9}})}                                   \\\bottomrule[1pt]
\end{tabular}
}
\end{table*}

\begin{table*}[p]
\centering
\caption{Full LTSF results on the benchmark datasets with Complex Models. Lower MSE and MAE indicate better forecasting accuracy. \textcolor{red}{\textbf{Red}} denote the best performance,  \textcolor{blue}{\underline{blue}} indicate the second-best results.}\label{tab. Full Complex Models}
\resizebox{0.75\textwidth}{!}{
\begin{tabular}{c|c|cc|cc|cc|cc|cc|cc|cc|cc|cc|cc}
\toprule[1pt]
                              &                      & \multicolumn{2}{c}{\method-t}                                & \multicolumn{2}{c|}{\method}                                   & \multicolumn{2}{c}{\texttt{TimeMixer}}                               & \multicolumn{2}{c}{\texttt{TiDE}}                                    & \multicolumn{2}{c}{\texttt{SCINet}} & \multicolumn{2}{c}{\texttt{TimesNet}}         & \multicolumn{2}{c}{\texttt{Autoformer}} & \multicolumn{2}{c}{\texttt{FEDFormer}} & \multicolumn{2}{c}{\texttt{PatchTST}}                                & \multicolumn{2}{c}{\texttt{iTransformer}}                            \\
                     & Pred\_len            & \multicolumn{1}{c}{MSE}      & \multicolumn{1}{c}{MAE}      & \multicolumn{1}{c}{MSE}      & \multicolumn{1}{c|}{MAE}      & \multicolumn{1}{c}{MSE}      & \multicolumn{1}{c}{MAE}                       & \multicolumn{1}{c}{MSE}      & \multicolumn{1}{c}{MAE}             & \multicolumn{1}{c}{MSE}      & \multicolumn{1}{c}{MAE}      & \multicolumn{1}{c}{MSE}      & \multicolumn{1}{c}{MAE}  & \multicolumn{1}{c}{MSE}      & \multicolumn{1}{c}{MAE}       & \multicolumn{1}{c}{MSE}      & \multicolumn{1}{c}{MAE}    & \multicolumn{1}{c}{MSE}      & \multicolumn{1}{c}{MAE}                         & \multicolumn{1}{c}{MSE}      & \multicolumn{1}{c}{MAE}                       \\\midrule
                              & 96                   & \textcolor{blue}{\underline{0.368}} & \textcolor{blue}{\underline{0.394}} & \textcolor{red}{\textbf{0.361}} & \textcolor{red}{\textbf{0.387}} & 0.397                        & 0.424                        & 0.378                        & 0.402                        & 0.397        & 0.418       & 0.453                        & 0.462 & 0.500          & 0.499         & 0.420         & 0.462         & 0.381                        & 0.410                        & 0.403                        & 0.426                        \\
                              & 192                  & \textcolor{red}{\textbf{0.406}} & \textcolor{blue}{\underline{0.418}} & \textcolor{blue}{\underline{0.409}} & \textcolor{red}{\textbf{0.408}} & 0.448                        & 0.458                        & 0.413                        & 0.424                        & 0.529        & 0.488       & 0.500                        & 0.494 & 0.516          & 0.519         & 0.474         & 0.498         & 0.420                        & 0.437                        & 0.435                        & 0.448                        \\
                              & 336                  & \textcolor{red}{\textbf{0.431}} & \textcolor{blue}{\underline{0.434}} & 0.445                        & \textcolor{red}{\textbf{0.428}} & 0.495                        & 0.491                        & \textcolor{blue}{\underline{0.436}} & 0.438                        & 0.476        & 0.470       & 0.515                        & 0.507 & 0.510          & 0.516         & 0.480         & 0.492         & 0.440                        & 0.452                        & 0.451                        & 0.462                        \\
\multirow{-4}{*}{\rotatebox{90}{ETTh1}}       & 720                  & 0.464                        & 0.472                        & \textcolor{blue}{\underline{0.457}} & \textcolor{red}{\textbf{0.461}} & 0.538                        & 0.528                        & \textcolor{red}{\textbf{0.456}} & \textcolor{blue}{\underline{0.467}} & 0.529        & 0.513       & 0.686                        & 0.593 & 0.650          & 0.607         & 0.547         & 0.540         & 0.496                        & 0.508                        & 0.570                        & 0.546                        \\\midrule
                              & 96                   & 0.283                        & 0.344                        & 0.283                        & \textcolor{red}{\textbf{0.329}} & 0.292                        & 0.357                        & \textcolor{red}{\textbf{0.277}} & \textcolor{blue}{\underline{0.339}} & 0.345        & 0.386       & 0.353                        & 0.405 & 0.412          & 0.462         & 0.392         & 0.450         & \textcolor{blue}{\underline{0.278}} & 0.340                        & 0.307                        & 0.363                        \\
                              & 192                  & 0.349                        & 0.388                        & 0.348                        & \textcolor{red}{\textbf{0.374}} & 0.362                        & 0.403                        & \textcolor{red}{\textbf{0.342}} & 0.383                        & 0.409        & 0.427       & 0.383                        & 0.422 & 0.425          & 0.471         & 0.429         & 0.474         & \textcolor{blue}{\underline{0.342}} & \textcolor{blue}{\underline{0.381}} & 0.375                        & 0.405                        \\
                              & 336                  & 0.379                        & 0.416                        & \textcolor{blue}{\underline{0.364}} & \textcolor{red}{\textbf{0.395}} & 0.378                        & 0.420                        & \textcolor{red}{\textbf{0.363}} & \textcolor{blue}{\underline{0.404}} & 0.407        & 0.439       & 0.399                        & 0.438 & 0.415          & 0.465         & 0.430         & 0.481         & 0.371                        & 0.408                        & 0.441                        & 0.446                        \\
\multirow{-4}{*}{\rotatebox{90}{ETTh2}}       & 720                  & 0.419                        & 0.448                        & 0.408                        & 0.446                        & \textcolor{red}{\textbf{0.379}} & \textcolor{red}{\textbf{0.426}} & \textcolor{blue}{\underline{0.389}} & \textcolor{blue}{\underline{0.431}} & 0.433        & 0.458       & 0.454                        & 0.473 & 0.499          & 0.522         & 0.498         & 0.516         & 0.395                        & 0.432                        & 0.459                        & 0.475                        \\\midrule
                              & 96                   & \textcolor{blue}{\underline{0.303}} & \textcolor{blue}{\underline{0.348}} & 0.305                        & \textcolor{red}{\textbf{0.341}} & 0.306                        & 0.360                        & 0.312                        & 0.354                        & 0.345        & 0.383       & 0.368                        & 0.395 & 0.576          & 0.504         & 0.382         & 0.435         & \textcolor{red}{\textbf{0.294}} & \textcolor{blue}{\underline{0.348}} & 0.314                        & 0.367                        \\
                              & 192                  & 0.337                        & \textcolor{blue}{\underline{0.368}} & \textcolor{red}{\textbf{0.320}} & \textcolor{red}{\textbf{0.351}} & 0.388                        & 0.405                        & 0.345                        & 0.373                        & 0.403        & 0.418       & 0.417                        & 0.417 & 0.571          & 0.509         & 0.398         & 0.440         & \textcolor{blue}{\underline{0.334}} & 0.372                        & 0.350                        & 0.388                        \\
                              & 336                  & \textcolor{blue}{\underline{0.369}} & \textcolor{blue}{\underline{0.385}} & \textcolor{red}{\textbf{0.353}} & \textcolor{red}{\textbf{0.378}} & 0.442                        & 0.440                        & 0.377                        & 0.392                        & 0.436        & 0.439       & 0.495                        & 0.459 & 0.579          & 0.511         & 0.445         & 0.464         & 0.371                        & 0.396                        & 0.380                        & 0.405                        \\
\multirow{-4}{*}{\rotatebox{90}{ETTm1}}       & 720                  & 0.422                        & \textcolor{blue}{\underline{0.414}} & \textcolor{red}{\textbf{0.407}} & \textcolor{red}{\textbf{0.405}} & 0.578                        & 0.506                        & 0.431                        & 0.421                        & 0.523        & 0.486       & 0.505                        & 0.479 & 0.551          & 0.497         & 0.503         & 0.492         & \textcolor{blue}{\underline{0.414}} & 0.426                        & 0.444                        & 0.444                        \\\midrule
                              & 96                   & \textcolor{red}{\textbf{0.165}} & \textcolor{blue}{\underline{0.254}} & 0.182                        & \textcolor{red}{\textbf{0.253}} & 0.182                        & 0.271                        & \textcolor{blue}{\underline{0.167}} & 0.255                        & 0.182        & 0.275       & 0.210                        & 0.290 & 0.281          & 0.354         & 0.278         & 0.354         & 0.169                        & 0.259                        & 0.181                        & 0.274                        \\
                              & 192                  & 0.224                        & \textcolor{blue}{\underline{0.293}} & 0.243                        & 0.304                        & 0.256                        & 0.319                        & \textcolor{red}{\textbf{0.220}} & \textcolor{red}{\textbf{0.291}} & 0.245        & 0.318       & 0.278                        & 0.334 & 0.305          & 0.366         & 0.311         & 0.375         & \textcolor{blue}{\underline{0.223}} & 0.298                        & 0.243                        & 0.315                        \\
                              & 336                  & \textcolor{blue}{\underline{0.280}} & \textcolor{blue}{\underline{0.331}} & 0.294                        & 0.341                        & 0.347                        & 0.380                        & \textcolor{red}{\textbf{0.276}} & \textcolor{red}{\textbf{0.329}} & 0.316        & 0.365       & 0.354                        & 0.375 & 0.343          & 0.386         & 0.349         & 0.393         & 0.282                        & 0.339                        & 0.289                        & 0.344                        \\
\multirow{-4}{*}{\rotatebox{90}{ETTm2}}       & 720                  & 0.369                        & \textcolor{blue}{\underline{0.387}} & \textcolor{red}{\textbf{0.339}} & 0.387                        & 0.461                        & 0.445                        & 0.364                        & \textcolor{red}{\textbf{0.385}} & 0.441        & 0.431       & 0.452                        & 0.434 & 0.432          & 0.450         & 0.434         & 0.447         & \textcolor{blue}{\underline{0.362}} & 0.389                        & 0.374                        & 0.396                        \\\midrule
                              & 96                   & \textcolor{blue}{\underline{0.156}} & \textcolor{blue}{\underline{0.207}} & 0.169                        & 0.223                        & 0.156                        & 0.215                        & 0.171                        & 0.224                        & 0.171        & 0.227       & 0.172                        & 0.227 & 0.296          & 0.359         & 0.298         & 0.366         & \textcolor{red}{\textbf{0.149}} & \textcolor{red}{\textbf{0.200}} & 0.164                        & 0.217                        \\
                              & 192                  & \textcolor{blue}{\underline{0.198}} & \textcolor{blue}{\underline{0.245}} & 0.214                        & 0.262                        & 0.202                        & 0.253                        & 0.214                        & 0.260                        & 0.223        & 0.271       & 0.233                        & 0.279 & 0.335          & 0.386         & 0.341         & 0.399         & \textcolor{red}{\textbf{0.193}} & \textcolor{red}{\textbf{0.241}} & 0.205                        & 0.253                        \\
                              & 336                  & \textcolor{blue}{\underline{0.250}} & \textcolor{blue}{\underline{0.286}} & 0.257                        & 0.293                        & 0.263                        & 0.299                        & 0.260                        & 0.294                        & 0.283        & 0.315       & 0.317                        & 0.336 & 0.388          & 0.420         & 0.397         & 0.440         & \textcolor{red}{\textbf{0.245}} & \textcolor{red}{\textbf{0.283}} & 0.253                        & 0.289                        \\
\multirow{-4}{*}{\rotatebox{90}{Weather}}     & 720                  & 0.321                        & \textcolor{blue}{\underline{0.336}} & 0.321                        & 0.340                        & 0.354                        & 0.354                        & 0.327                        & 0.342                        & 0.359        & 0.362       & 0.377                        & 0.376 & 0.401          & 0.425         & 0.427         & 0.447         & \textcolor{red}{\textbf{0.316}} & \textcolor{red}{\textbf{0.334}} & \textcolor{blue}{\underline{0.318}} & \textcolor{blue}{\underline{0.336}} \\\midrule
                              & 96                   & 0.135                        & 0.231                        & \textcolor{red}{\textbf{0.127}} & \textcolor{red}{\textbf{0.221}} & 0.151                        & 0.256                        & 0.136                        & 0.232                        & 0.193        & 0.306       & 0.186                        & 0.291 & 0.273          & 0.374         & 0.219         & 0.333         & 0.133                        & \textcolor{blue}{\underline{0.229}} & \textcolor{blue}{\underline{0.133}} & 0.229                        \\
                              & 192                  & 0.151                        & 0.245                        & \textcolor{red}{\textbf{0.143}} & \textcolor{red}{\textbf{0.234}} & 0.167                        & 0.272                        & \textcolor{blue}{\underline{0.150}} & \textcolor{blue}{\underline{0.244}} & 0.198        & 0.307       & 0.194                        & 0.294 & 0.256          & 0.358         & 0.228         & 0.341         & \textcolor{blue}{\underline{0.150}} & 0.246                        & 0.152                        & 0.250                        \\
                              & 336                  & 0.167                        & \textcolor{blue}{\underline{0.261}} & \textcolor{red}{\textbf{0.158}} & \textcolor{red}{\textbf{0.248}} & 0.196                        & 0.297                        & \textcolor{blue}{\underline{0.167}} & \textcolor{blue}{\underline{0.261}} & 0.221        & 0.328       & 0.239                        & 0.330 & 0.267          & 0.368         & 0.228         & 0.341         & 0.167                        & 0.263                        & \textcolor{blue}{\underline{0.167}} & 0.266                        \\
\multirow{-4}{*}{\rotatebox{90}{Electricity}} & 720                  & 0.207                        & 0.295                        & \textcolor{blue}{\underline{0.198}} & \textcolor{red}{\textbf{0.283}} & 0.225                        & 0.320                        & 0.206                        & 0.294                        & 0.257        & 0.353       & 0.236                        & 0.328 & 0.307          & 0.399         & 0.248         & 0.357         & 0.204                        & 0.295                        & \textcolor{red}{\textbf{0.192}} & \textcolor{blue}{\underline{0.288}} \\\midrule
                              & 96                   & 0.395                        & 0.275                        & \textcolor{red}{\textbf{0.353}} & \textcolor{red}{\textbf{0.231}} & 0.416                        & 0.310                        & 0.407                        & 0.297                        & 0.492        & 0.394       & 0.599                        & 0.325 & 0.631          & 0.393         & 0.591         & 0.368         & 0.369                        & \textcolor{blue}{\underline{0.257}} & \textcolor{blue}{\underline{0.363}} & 0.262                        \\
                              & 192                  & 0.408                        & 0.281                        & \textcolor{red}{\textbf{0.364}} & \textcolor{red}{\textbf{0.243}} & 0.435                        & 0.318                        & 0.423                        & 0.305                        & 0.496        & 0.393       & 0.618                        & 0.333 & 0.684          & 0.424         & 0.600         & 0.366         & \textcolor{blue}{\underline{0.384}} & \textcolor{blue}{\underline{0.264}} & 0.390                        & 0.281                        \\
                              & 336                  & 0.418                        & 0.285                        & \textcolor{red}{\textbf{0.375}} & \textcolor{red}{\textbf{0.248}} & 0.439                        & 0.320                        & 0.433                        & 0.311                        & 0.509        & 0.398       & 0.617                        & 0.336 & 0.680          & 0.410         & 0.614         & 0.374         & \textcolor{blue}{\underline{0.396}} & \textcolor{blue}{\underline{0.271}} & 0.401                        & 0.287                        \\
\multirow{-4}{*}{\rotatebox{90}{Traffic}}     & 720                  & 0.454                        & 0.305                        & \textcolor{red}{\textbf{0.421}} & \textcolor{red}{\textbf{0.271}} & 0.463                        & 0.330                        & 0.480                        & 0.341                        & 0.545        & 0.413       & 0.656                        & 0.347 & 0.671          & 0.403         & 0.633         & 0.385         & \textcolor{blue}{\underline{0.436}} & \textcolor{blue}{\underline{0.294}} & 0.440                        & 0.307                        \\\midrule
                              & 96                   & 0.206                        & 0.257                        & \textcolor{blue}{\underline{0.183}} & \textcolor{blue}{\underline{0.238}} & 0.215                        & 0.268                        & 0.210                        & 0.258                        & 0.210        & 0.295       & 0.191                        & 0.270 & 0.679          & 0.604         & 0.351         & 0.446         & \textcolor{red}{\textbf{0.168}} & \textcolor{red}{\textbf{0.237}} & 0.189                        & 0.247                        \\
                              & 192                  & 0.229                        & 0.271                        & 0.213                        & \textcolor{red}{\textbf{0.243}} & 0.213                        & 0.274                        & 0.231                        & 0.271                        & 0.228        & 0.314       & 0.214                        & 0.288 & 0.981          & 0.760         & 0.341         & 0.432         & \textcolor{red}{\textbf{0.184}} & \textcolor{blue}{\underline{0.246}} & \textcolor{blue}{\underline{0.206}} & 0.272                        \\
                              & 336                  & 0.242                        & 0.281                        & 0.228                        & \textcolor{red}{\textbf{0.243}} & 0.240                        & 0.287                        & 0.247                        & 0.277                        & 0.238        & 0.318       & \textcolor{blue}{\underline{0.216}} & 0.285 & 0.842          & 0.690         & 0.300         & 0.380         & \textcolor{red}{\textbf{0.192}} & \textcolor{blue}{\underline{0.253}} & 0.220                        & 0.285                        \\
\multirow{-4}{*}{\rotatebox{90}{Solar}}       & 720                  & 0.250                        & 0.285                        & 0.228                        & \textcolor{red}{\textbf{0.254}} & 0.232                        & 0.289                        & 0.258                        & 0.293                        & 0.234        & 0.314       & 0.244                        & 0.303 & 0.888          & 0.716         & 0.328         & 0.400         & \textcolor{red}{\textbf{0.204}} & \textcolor{blue}{\underline{0.263}} & \textcolor{blue}{\underline{0.211}} & 0.272                        \\\midrule
\multicolumn{2}{c}{Wins}                             & \multicolumn{2}{c}{\textcolor{red}{\textbf{3}} ( \textcolor{blue}{\underline{23}})}                                   & \multicolumn{2}{c}{\textcolor{red}{\textbf{35}} ( \textcolor{blue}{\underline{6}})}                                   & \multicolumn{2}{c}{\textcolor{red}{\textbf{2}} ( \textcolor{blue}{\underline{0}})}                                    & \multicolumn{2}{c}{\textcolor{red}{\textbf{9}} ( \textcolor{blue}{\underline{11}})}                                   & \multicolumn{2}{c}{\textcolor{red}{\textbf{0}} ( \textcolor{blue}{\underline{0}})}   & \multicolumn{2}{c}{\textcolor{red}{\textbf{0}} ( \textcolor{blue}{\underline{1}})}             & \multicolumn{2}{c}{\textcolor{red}{\textbf{0}} ( \textcolor{blue}{\underline{0}})}       & \multicolumn{2}{c}{\textcolor{red}{\textbf{0}} ( \textcolor{blue}{\underline{0}})}      & \multicolumn{2}{c}{\textcolor{red}{\textbf{14}} ( \textcolor{blue}{\underline{20}})}                                  & \multicolumn{2}{c}{\textcolor{red}{\textbf{1}} ( \textcolor{blue}{\underline{8}})}                                    \\ \bottomrule[1pt]
\end{tabular}
}
\end{table*}

\subsection{Full Ablation Results}\label{app.AblationResults}
In Table~\red{\ref{tab. ablation}}, we present the ablation analysis of \method(-t), where the MSE and MAE values are averaged over different prediction horizons.
Table~\red{\ref{tab. Full Ablation}} further provides the complete results covering all prediction lengths $H \in \{96,192,336,720\}$ for comprehensive reference.
We observe that removing the patch-wise variety, the distance-attenuating local attention, or the back-cast mechanism at different horizons leads to a clear performance degradation.
This further demonstrates the indispensable role of these components in \method(-t), as they collectively contribute significantly to the overall forecasting performance.

\begin{table*}[htb]
\centering
\caption{Full Ablation analysis results of \method (-t). Lower MSE and MAE are better.}\label{tab. Full Ablation}
\resizebox{0.85\textwidth}{!}{
\begin{tabular}{l|c|cc|cc|cc|cc|cc|cc|cc}
\toprule[1pt]
           &        & \multicolumn{2}{c}{ETTh1} & \multicolumn{2}{c}{ETTh2} & \multicolumn{2}{c}{ETTm1} & \multicolumn{2}{c}{ETTm2} & \multicolumn{2}{c}{Weather} & \multicolumn{2}{c}{Electricity} & \multicolumn{2}{c}{Traffic} \\
\textbf{Method}     & Length & MSE         & MAE         & MSE         & MAE         & MSE         & MAE         & MSE         & MAE         & MSE          & MAE          & MSE            & MAE            & MSE          & MAE          \\\midrule
SVTime             & 96     & 0.361       & 0.387       & 0.283       & 0.329       & 0.305       & 0.341       & 0.182       & 0.253       & 0.169        & 0.223        & 0.127          & 0.221          & 0.353        & 0.231        \\
                   & 192    & 0.409       & 0.408       & 0.348       & 0.374       & 0.320       & 0.351       & 0.243       & 0.304       & 0.214        & 0.262        & 0.143          & 0.234          & 0.364        & 0.243        \\
                   & 336    & 0.445       & 0.428       & 0.364       & 0.395       & 0.353       & 0.378       & 0.294       & 0.341       & 0.257        & 0.293        & 0.158          & 0.248          & 0.375        & 0.248        \\
                   & 720    & 0.457       & 0.461       & 0.408       & 0.446       & 0.407       & 0.405       & 0.339       & 0.387       & 0.321        & 0.340        & 0.198          & 0.283          & 0.421        & 0.271        \\
                   & Avg.   & 0.418       & 0.421       & 0.351       & 0.386       & 0.346       & 0.369       & 0.265       & 0.321       & 0.240        & 0.280        & 0.157          & 0.247          & 0.378        & 0.248        \\\midrule
(a) - IB2 & 96     & 0.371       & 0.394       & 0.274       & 0.338       & 0.301       & 0.346       & 0.164       & 0.254       & 0.17         & 0.221        & 0.136          & 0.231          & 0.414        & 0.301        \\
                   & 192    & 0.405       & 0.416       & 0.335       & 0.378       & 0.339       & 0.368       & 0.22        & 0.291       & 0.214        & 0.26         & 0.151          & 0.244          & 0.489        & 0.336        \\
                   & 336    & 0.437       & 0.436       & 0.367       & 0.406       & 0.372       & 0.387       & 0.273       & 0.33        & 0.26         & 0.294        & 0.166          & 0.26           & 0.513        & 0.368        \\
                   & 720    & 0.467       & 0.468       & 0.391       & 0.432       & 0.43        & 0.419       & 0.357       & 0.382       & 0.326        & 0.341        & 0.205          & 0.292          & 0.523        & 0.371        \\
                   & Avg.   & 0.420       & 0.429       & 0.342       & 0.389       & 0.361       & 0.380       & 0.254       & 0.314       & 0.243        & 0.279        & 0.165          & 0.257          & 0.485        & 0.344        \\\midrule
(b) - Backcast  & 96     & 0.367       & 0.389       & 0.309       & 0.352       & 0.298       & 0.344       & 0.172       & 0.262       & 0.198        & 0.259        & 0.186          & 0.270          & 0.454        & 0.254        \\
                   & 192    & 0.412       & 0.416       & 0.360       & 0.390       & 0.333       & 0.363       & 0.230       & 0.299       & 0.243        & 0.292        & 0.202          & 0.283          & 0.519        & 0.281        \\
                   & 336    & 0.427       & 0.431       & 0.376       & 0.409       & 0.506       & 0.494       & 0.331       & 0.375       & 0.289        & 0.322        & 0.217          & 0.298          & 0.534        & 0.289        \\
                   & 720    & 0.439       & 0.450       & 0.433       & 0.457       & 0.544       & 0.511       & 0.416       & 0.424       & 0.323        & 0.333        & 0.253          & 0.327          & 0.566        & 0.304        \\
                   & Avg.   & 0.411       & 0.421       & 0.369       & 0.402       & 0.420       & 0.428       & 0.287       & 0.340       & 0.263        & 0.301        & 0.214          & 0.295          & 0.518        & 0.282        \\\hline
SVTime-t           & 96     & 0.368       & 0.394       & 0.283       & 0.344       & 0.303       & 0.348       & 0.165       & 0.254       & 0.156        & 0.207        & 0.135          & 0.231          & 0.395        & 0.275        \\
                   & 192    & 0.406       & 0.418       & 0.349       & 0.388       & 0.337       & 0.368       & 0.224       & 0.293       & 0.198        & 0.245        & 0.151          & 0.245          & 0.408        & 0.281        \\
                   & 336    & 0.431       & 0.434       & 0.379       & 0.416       & 0.369       & 0.385       & 0.280       & 0.331       & 0.250        & 0.286        & 0.167          & 0.261          & 0.418        & 0.285        \\
                   & 720    & 0.464       & 0.472       & 0.419       & 0.448       & 0.422       & 0.414       & 0.369       & 0.387       & 0.321        & 0.336        & 0.207          & 0.295          & 0.454        & 0.305        \\
                   & Avg.   & 0.417       & 0.430       & 0.357       & 0.399       & 0.358       & 0.379       & 0.259       & 0.316       & 0.231        & 0.269        & 0.165          & 0.258          & 0.419        & 0.286        \\\midrule
(c) - IB2 & 96     & 0.385       & 0.405       & 0.301       & 0.369       & 0.318       & 0.352       & 0.218       & 0.289       & 0.249        & 0.286        & 0.143          & 0.237          & 0.404        & 0.275        \\
                   & 192    & 0.405       & 0.410       & 0.352       & 0.395       & 0.340       & 0.364       & 0.258       & 0.316       & 0.270        & 0.300        & 0.155          & 0.247          & 0.414        & 0.278        \\
                   & 336    & 0.432       & 0.426       & 0.382       & 0.425       & 0.368       & 0.381       & 0.301       & 0.344       & 0.301        & 0.320        & 0.170          & 0.263          & 0.423        & 0.283        \\
                   & 720    & 0.468       & 0.472       & 0.431       & 0.456       & 0.420       & 0.412       & 0.380       & 0.394       & 0.350        & 0.354        & 0.210          & 0.296          & 0.459        & 0.302        \\
                   & Avg.   & 0.422       & 0.428       & 0.366       & 0.411       & 0.361       & 0.377       & 0.289       & 0.336       & 0.293        & 0.315        & 0.170          & 0.261          & 0.425        & 0.284        \\\midrule
(d) - IB3   & 96     & 0.385       & 0.415       & 0.303       & 0.367       & 0.318       & 0.353       & 0.218       & 0.290       & 0.170        & 0.222        & 0.143          & 0.237          & 0.404        & 0.274        \\
                   & 192    & 0.424       & 0.437       & 0.362       & 0.389       & 0.341       & 0.365       & 0.258       & 0.316       & 0.215        & 0.260        & 0.155          & 0.247          & 0.414        & 0.277        \\
                   & 336    & 0.449       & 0.441       & 0.380       & 0.408       & 0.368       & 0.382       & 0.301       & 0.344       & 0.260        & 0.294        & 0.170          & 0.263          & 0.422        & 0.282        \\
                   & 720    & 0.452       & 0.469       & 0.393       & 0.432       & 0.421       & 0.412       & 0.380       & 0.393       & 0.326        & 0.342        & 0.209          & 0.296          & 0.458        & 0.299        \\
                   & Avg.   & 0.428       & 0.440       & 0.359       & 0.399       & 0.362       & 0.378       & 0.289       & 0.336       & 0.243        & 0.280        & 0.169          & 0.260          & 0.425        & 0.283        \\\midrule
(e) - Backcast  & 96     & 0.435       & 0.463       & 0.335       & 0.377       & 0.545       & 0.434       & 0.248       & 0.314       & 0.25         & 0.29         & 0.18           & 0.265          & 0.564        & 0.371        \\
                   & 192    & 0.471       & 0.482       & 0.37        & 0.398       & 0.553       & 0.441       & 0.282       & 0.335       & 0.273        & 0.304        & 0.199          & 0.282          & 0.623        & 0.386        \\
                   & 336    & 0.513       & 0.543       & 0.375       & 0.407       & 0.568       & 0.451       & 0.321       & 0.359       & 0.303        & 0.324        & 0.215          & 0.296          & 0.634        & 0.389        \\
                   & 720    & 0.532       & 0.552       & 0.404       & 0.433       & 0.593       & 0.471       & 0.401       & 0.408       & 0.352        & 0.359        & 0.252          & 0.329          & 0.668        & 0.404        \\
                   & Avg.   & 0.488       & 0.510       & 0.371       & 0.404       & 0.565       & 0.449       & 0.313       & 0.354       & 0.295        & 0.319        & 0.212          & 0.293          & 0.622        & 0.387       \\
\bottomrule[1pt]
\end{tabular}
}
\end{table*}

\subsection{Comparison with Pre-trained Models}\label{app. Pre-trained Models}
Table~\red{\ref{tab.Pretrained model}}, as a complement to Fig.~\red{\ref{fig.large_model}}, provides a more extensive comparison between \method and various large pre-trained models on general time series datasets. CALF\cite{liu2025calf}, GPT4TS\cite{zhou2023one} and TimeLLM\cite{jin2024time} are based on language pretrained models; VisionTS\cite{chen2025visionts} leverages vision pretrained models; TimeVLM\cite{zhong2025time} uses vision and language pretrained models; and LightGTS\cite{wang2025lightgts} is pretrained purely on time series data.
Our model achieves the best performance on four out of seven datasets, and on the remaining three datasets, the performance gap does not exceed 10\%. Considering the substantial differences in parameter scale and inference speed between \method and these large pre-trained models, such comparable results demonstrate that directly training a lightweight model on the target dataset—without costly pre-training and fine-tuning—can still be highly effective and valuable.

\begin{table*}[htb]
\centering
\caption{The comparison with Pretrained time series models. \textcolor{red}{Red} denotes performance superior to \method.}\label{tab.Pretrained model}
\resizebox{0.85\textwidth}{!}{
\begin{tabular}{l|cc|cc|cc|cc|cc|cc|cc}
\toprule[1pt]
& \multicolumn{2}{c}{ETTh1}                               & \multicolumn{2}{c}{ETTh2}                                & \multicolumn{2}{c}{ETTm1} & \multicolumn{2}{c}{ETTm2} & \multicolumn{2}{c}{Weather}                             & \multicolumn{2}{c}{Electricity} & \multicolumn{2}{c}{Traffic} \\
Model                & MSE                        & MAE                        & MSE                         & MAE                        & MSE         & MAE         & MSE         & MAE         & MSE                        & MAE                        & MSE            & MAE            & MSE          & MAE          \\\midrule

\method                   & 0.418                      & 0.421                      & 0.351                       & 0.386                      & \textbf{0.346}       & \textbf{0.369}       & \textbf{0.265}       & \textbf{0.321}       & 0.240                      & 0.280                      & \textbf{0.157}          & \textbf{0.247}          & \textbf{0.378}        & \textbf{0.248}       \\ \midrule
 \texttt{CALF} (2024)                                                & 0.432                      & 0.431                      & 0.333                       & \textbf{0.369}                      & 0.368       & 0.385       & 0.355       & 0.365       & 0.280                      & 0.304                      & 0.176          & 0.266          & 0.421        & 0.274        \\
Improvement   & +3\%                       & +2\%                       & \textcolor{red}{ -5\%}  & \textcolor{red}{ -4\%} & +6\%        & +4\%        & +34\%       & +14\%       & +17\%                      & +9\%                       & +12\%          &+8\%           & +11\%        & +10\%        \\\midrule
\texttt{GPT4TS} (2023)                                                & 0.418                      & 0.421                      & 0.336                       & 0.373                      & 0.350       & 0.381       & 0.323       & 0.350       & 0.251                      & 0.288                      & 0.170          & 0.263          & 0.421        & 0.274        \\
Improvement & 0\%                        & +0\%                        & \textcolor{red}{ -4\%}  & \textcolor{red}{ -3\%} & +1\%        & +3\%        & +22\%       & +9\%        & +5\%                       & +3\%                       & +8\%           & +7\%           & +11\%        & +10\%        \\\midrule
\texttt{TimeLLM} (2024)                                           & 0.418                      & 0.432                      & 0.346                       & 0.384                      & 0.346       & 0.381       & 0.318       & 0.349       & 0.265                      & 0.299                      & 0.165          & 0.259          & 0.422        & 0.281        \\
Improvement   & 0\%                        & +3\%                       & \textcolor{red}{ -1\%}  & \textcolor{red}{ -1\%} & 0\%         & +3\%        & +20\%       & +9\%        & +10\%                      & +7\%                       & +5\%           & +5\%           & +11\%        & +13\%        \\\midrule
 \texttt{VisionTS} (2025)                                            & 0.409                      & \textbf{0.417}                      & 0.359                       & 0.391                      & 0.345       & 0.373       & 0.269       & 0.328       & \textbf{0.224}                      & \textbf{0.257}                      & 0.161          & 0.253          & 0.387        & 0.255        \\
Improvement  & \textcolor{red}{ -2\%} & \textcolor{red}{ -1\%} & +2\%                        & +1\%                       & 0\%         & +1\%        & +2\%        & +2\%        & \textcolor{red}{ -7\%} & \textcolor{red}{ -8\%} & +3\%           & +2\%           & +2\%         & +3\%         \\\midrule
\texttt{TimeVLM} (2025)                               & 0.405                      & 0.420                      & \textbf{0.317}                       & 0.371                      & 0.349       & 0.387       & 0.309       & 0.348       & 0.247                      & 0.292                      & 0.172          & 0.272          & 0.419        & 0.304        \\
Improvement & \textcolor{red}{ -3\%} & 0\%                        & \textcolor{red}{ -10\%} & \textcolor{red}{ -4\%} & +1\%        & +5\%        & +17\%       & +8\%        & +3\%                       & +5\%                       & +10\%          & +10\%          & +11\%        & +22\%        \\\midrule
\texttt{LightGTS} (2025)                                   & \textbf{0.394}                      & 0.419                      & 0.323                       & 0.383                      & 0.346       & 0.379       & 0.393       & 0.383       & 0.257                      & 0.283                      & 0.187          & 0.279          & 0.439        & 0.329        \\
Improvement& \textcolor{red}{ -6\%} & \textcolor{red}{ -1\%} & \textcolor{red}{ -8\%}  & \textcolor{red}{ -1\%} & 0\%         & +3\%        & +49\%       & +19\%       & +7\%                       & +1\%                       & +19\%          & +13\%          & +16\%        & +32\%        \\
\bottomrule[1pt]
\end{tabular}}
\end{table*}

\end{document}